\documentclass[conference]{IEEEtran}
\pdfoutput=1
\usepackage{times}
\usepackage{caption}
\usepackage{graphicx}
\let\labelindent\relax 
\usepackage{enumitem}
\usepackage[numbers]{natbib}
\usepackage{multicol}
\usepackage[bookmarks=true]{hyperref}
\usepackage{multirow}
\usepackage{booktabs}
\usepackage[dvipsnames]{xcolor}
\usepackage{amsmath}
\usepackage{cleveref}
\usepackage{xspace}
\usepackage[OT1]{fontenc} 
\usepackage{svg}
\usepackage{algorithm}
\usepackage[noend]{algorithmic}
\usepackage[breakable]{tcolorbox}
\usepackage{subcaption}

\definecolor{mypink}{RGB}{219, 48, 122}

\newcommand{\our}{\textsc{RHINO}\xspace}

\begin{document}

\title{RHINO: Learning Real-Time Humanoid-Human-Object Interaction\\ from Human Demonstrations}

\author{\authorblockN{Jingxiao Chen\authorrefmark{1},
Xinyao Li\authorrefmark{1},
Jiahang Cao\authorrefmark{1},
Zhengbang Zhu,
Wentao Dong, \\
Minghuan Liu\authorrefmark{2},
Ying Wen,
Yong Yu,
Liqing Zhang,
Weinan Zhang}
\authorblockA{\authorrefmark{1}Equal Contribution~~\authorrefmark{2}Project Lead}
 \authorblockA{Shanghai Jiao Tong University}
 \authorblockA{ \textcolor{RubineRed}{\href{https://humanoid-interaction.github.io}{humanoid-interaction.github.io}}}
 }


\twocolumn[{%
\renewcommand\twocolumn[1][]{#1}%
\maketitle
\begin{center}
    \vspace{-6mm}
    \includegraphics[width=0.95\linewidth]{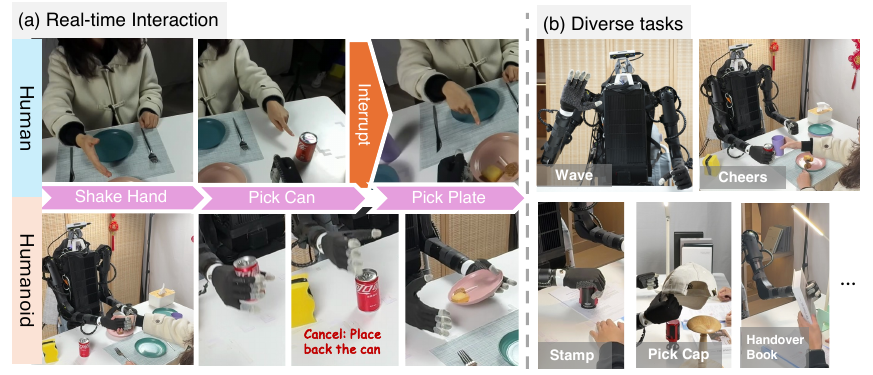}
    
\end{center}
\vspace{-10pt}
\captionof{figure}{
\textbf{\our has the capabilities of real-time interaction on diverse tasks.}
(a) \our enables real-time humanoid-human-object interaction, allowing seamless task interruption and dynamic switching during operation.
(b) The system demonstrates diverse capabilities, including waving, cheering, stamping, object pickup, handovers, and more.
}
\label{fig:teaser}
}]
\begin{abstract}
    Humanoid robots have shown success in locomotion and manipulation. Despite these basic abilities, humanoids are still required to quickly understand human instructions and react based on human interaction signals to become valuable assistants in human daily life.
    Unfortunately, most existing works only focus on multi-stage interactions, treating each task separately, and neglecting real-time feedback.
    In this work, we aim to empower humanoid robots with real-time reaction abilities to achieve various tasks, allowing human to interrupt robots at any time, and making robots respond to humans immediately.
    To support such abilities, we propose a general humanoid-human-object interaction framework, named \our, i.e., Real-time Humanoid-human Interaction and Object manipulation.
    \our provides a unified view of reactive motion, instruction-based manipulation, and safety concerns, over multiple human signal modalities, such as languages, images, and motions.
    \our is a hierarchical learning framework, enabling humanoids to learn reaction skills from human-human-object demonstrations and teleoperation data.
    In particular, it decouples the interaction process into two levels: 1) a high-level planner inferring human intentions from real-time human behaviors; and 2) a low-level controller achieving reactive motion behaviors and object manipulation skills based on the predicted intentions.
    We evaluate the proposed framework on a real humanoid robot and demonstrate its effectiveness, flexibility, and safety in various scenarios.
\end{abstract}

\IEEEpeerreviewmaketitle

\section{Introduction}
Humanoid robots are increasingly being explored to perform tasks in diverse environments~\citep{agravante2019human,johnson2015team,Kheddar2019}. Their human-like morphology provides a potential for acting with human-like dexterity, making them ideal for general-purpose daily-life human assistants.
However, most recent progresses only focus on learning basic abilities such as locomotion~\cite{HumanoidTransformer2023}, object manipulation~\cite{cheng2024tv}, and expressive motion~\cite{cheng2024express}.

Considering how we as humans react to our friends, a practically helpful humanoid assistant should possess three fundamental capabilities:
1) skill proficiency, equipped with diverse and essential skills to achieve various tasks;
2) intention recognition, capable of discerning human intentions, from either motion or language; and
3) instant feedback, able to respond in real-time with feasible actions. 
Nonetheless, most studies on human-robot interaction only focus on only one or two of these aspects. 
For instance, a significant body of work on human-robot interaction focuses on
object handover~\citep{strabalaSeamlessHumanRobotHandovers2013, tulbureFastPerceptionHumanRobot2024}, or interactive motion generation~\citep{prasadMoVEIntMixtureVariational2024, butepageImitatingGeneratingDeep2019, liangInterGenDiffusionbasedMultihuman2024, liuPhysReactionPhysicallyPlausible2024, mascaroRobotInteractionBehavior2024}, lacking the ability to switch between different tasks in real-time.
Some others focus on recognizing human intentions~\citep{duarteActionAnticipationReading2018, enanRoboticDetectionHumanComprehensible2022, fangEgoPAT3Dv2Predicting3D2024, mascaro2023hoi4abot, scherfAreYouSure2024}, which simplify the diversity of reaction and treat the interaction as an alternated two-stage process. The robot cannot be interrupted once a task is in progress, and further human commands can only be executed after the completion of the robot's current task.
Many recent works have attempted to combine the ability of general foundation models to enable robots to understand the complexity of human interactions~\citep{tanneberg2024help, wuHumanObjectInteractionHumanLevel2024}, but they often suffer from high latency and are not suitable for real-time interaction tasks.
These limitations hinder robots from rapid interventions and robust, multi-step interactions in human-centered tasks. Therefore, a unified framework that masters human-robot interaction with real-time intention recognition and various skills is urgently needed to tackle the above challenges.


To achieve this goal, we propose \our, a hierarchical learning framework for \underline{R}eactive \underline{H}umanoid-human \underline{IN}teraction and \underline{O}bject Manipulation.
\our decouples the interaction process into two levels: a high-level planner that infers human intentions from real-time human behaviors, and a low-level controller that achieves reactive motion behaviors and object manipulation skills based on predicted intentions. The high-level planner updates at high frequency, and the low-level controller is designed to be interruptable, enabling it to react to high-level commands at any time.
To ensure the scalability of \our across a wide range of skills, we design a pipeline for learning the interactions from human-object-human demonstration and teleoperation data, which can be easily extended to different tasks and scenarios.
We implement \our on a real humanoid robot and demonstrate its effectiveness, flexibility, and safety in various scenarios (see \Cref{fig:teaser}). 
Although this work only focuses on the upper body of a humanoid including the head, arms, and hands, it has the potential to be extended to whole-body humanoid interaction with a unified humanoid controller, and finally brings robots, especially humanoids, closer to our daily lives.

Our main contributions are in the following aspects:
\begin{itemize}[leftmargin=*]
    \item We propose the first humanoid learning architecture that seamlessly integrates intention recognition with real-time human-object-humanoid interaction skills, enabling the robot to respond to human instruction and switch between different tasks immediately. 
    \item We design a pipeline for learning the interactions from human demonstrations, which can easily scale to different tasks and scenarios.
    \item We implement \our on the Unitree H1 humanoid robot and demonstrate its effectiveness, flexibility, and safety in 2 scenarios with over 20 tasks, and open-source the code and datasets to facilitate future research.
\end{itemize}

    

\begin{figure*}[ht]
    \centering
    \includegraphics[width=\linewidth]{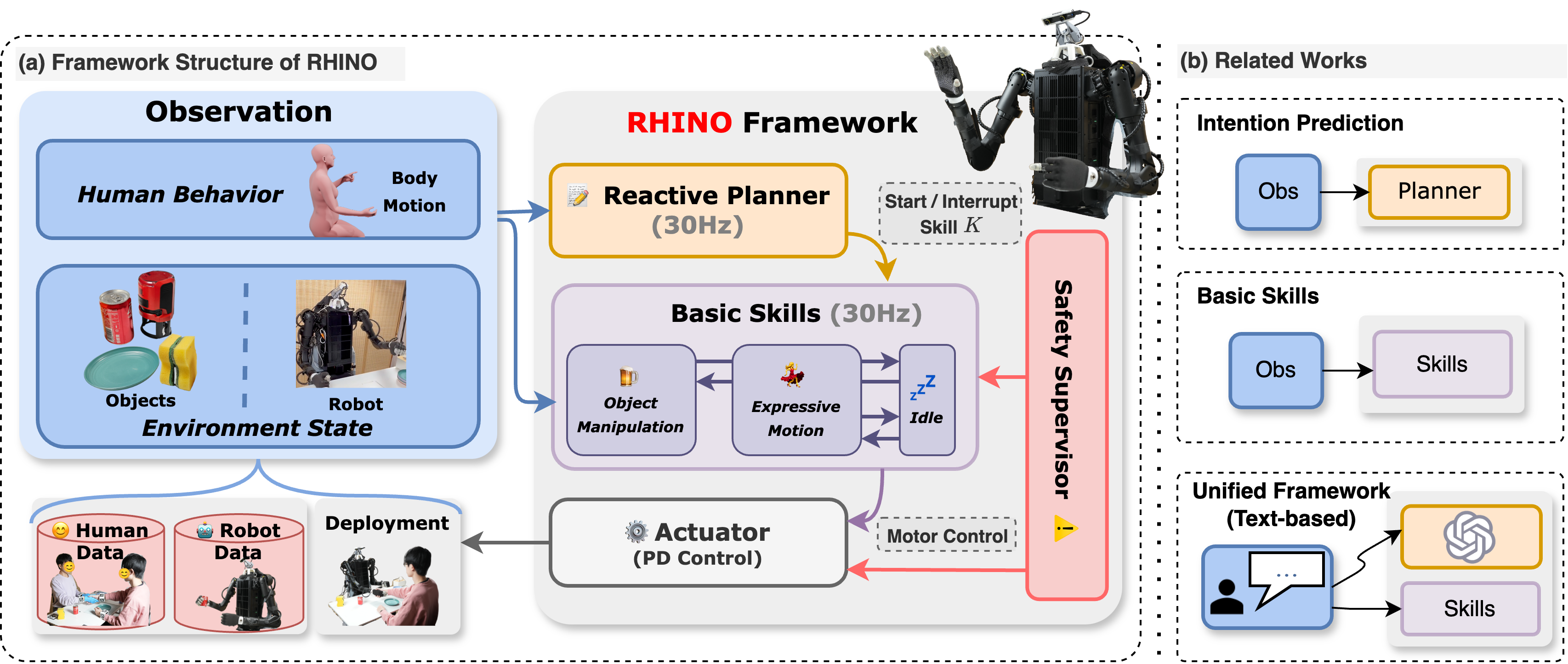}
    \caption{
        \textbf{Components of the \our framework and related works.}
        (a) Illustration of the \our framework, including the reactive planner, motion generation skills, object interaction skills and safety supervisor. 
    (b) Classification of related works based on the components of the \our framework.
    }
    \label{fig:formulation}
\end{figure*} 
\section{Related Works}
Recent progress in building a human assistant robot can be divided into three categories: 1) recognition of human intention,n, 2) basic skills, and 3) unified interaction framework, as shown in \Cref{fig:formulation}(b).
We summarize related works in each category and highlight the differences between our work.

\subsection{Human Intention Recognition.}
Humanoid robots need to estimate the human physical and mental states to provide appropriate assistance~\cite{vianelloHumanHumanoidInteractionCooperation2021}.
More specifically, many signals can be used to infer human intentions, such as whole-body motion~\cite{tulbureFastPerceptionHumanRobot2024, yangReactiveHumantoRobotHandovers2021}, forces~\cite{agravante2019human}, gaze~\cite{duarteActionAnticipationReading2018,scherfAreYouSure2024}, and language~\cite{tanneberg2024help}.
Object information in the environment also plays an important role in predicting human intention by combining it with human motion.
Human-object interaction, such as pointing gestures~\cite{fangEgoPAT3Dv2Predicting3D2024} and grabbing objects~\cite{mascaro2023hoi4abot}, provides a broader semantic space for human actions.
Most works on human intention recognition treat the interaction as a two-stage process, where the robot first predicts the human intention and then executes the task. This design simplifies the diversity of reactions and neglects the real-time reaction ability of the robot. Our work aims to react to human signals in real time, enabling the downstream tasks to be interrupted at any time.

\subsection{Basic Skills}
\noindent\textbf{Interactive motion synthesis.}
In human-robot interaction (HRI), learning to generate interactive and expressive motions, such as shaking hands and waving, are fundamental skills. 
The human-like morphology of humanoid robots provides a unique opportunity to learn natural motion from retargeted human motion data~\citep{fuHumanPlusHumanoidShadowing2024}. Human motion data can be collected from motion capture systems or network videos. Compared to collecting robot motion data, it has a lower cost and higher scalability.
Recent works~\cite{liangInterGenDiffusionbasedMultihuman2024, xuInterXVersatileHumanHuman2023} collect multi-human motion data, capturing real-time interaction and reaction between humans. 
Building on this, studies encode social scenes \cite{mascaroRobotInteractionBehavior2024}, simulate reactions \cite{liuPhysReactionPhysicallyPlausible2024}, or deploy interaction models on robots \cite{prasadMoVEIntMixtureVariational2024}. Our work focuses on learning interactive motion from human-human-object interaction data.

\noindent\textbf{Object manipulation.}
The ability to manipulate objects is another fundamental skill for a humanoid assistant robot, which requires more precise control of the robot's end-effector.
Limited by the dexterity of the robot, especially the degree of freedom of our humanoid robot's arm and hand, imitating learning from real-world teleoperation data~\cite{chi2023diffusion,pari2021surprising,mandlekar2021matters} is a more practical way to ensure success, compared to learning from human data~\cite{wang2023mimicplay, qin2024anyteleopgeneralvisionbaseddexterous, zhu2024vision}.
Open-Television~\cite{cheng2024tv} developed a teleoperation system with a VR device to control the arms and neck of humanoid robots and show the effectiveness and efficiency of learning manipulation skills with ACT~\cite{zhao2023learning} policy.
Our work learns the manipulation skills based on the teleoperation data.

\subsection{Unified Interaction Framework}
Recent works have attempted to leverage the capacity of general foundation models, such as large language models (LLMs) or vison-language models (VLMs), to enable robots to understand human intention in the format of text-based instructions~\cite{tanneberg2024help}. 
However, such interaction is often high-latency and not suitable for real-time environments, limiting the potential for natural and effective human-robot collaboration, particularly in scenarios that require immediate response or adaptation to changing human needs. 
\citet{asfourARMAR6HighPerformanceHumanoid2019} designed rules of the real-time human-robot interaction, which is hard to scale up.
\citet{cardenas2024xbg} tries to learn an end-to-end model by imitation to achieve real-time interaction with 5 different tasks. Limited by the sample efficiency, this end-to-end paradigm makes it difficult to scale to more tasks.

Our framework decouples the interaction process and enables each module to model the interaction with different observation spaces, which is more sample-efficient and scalable. We also deploy the framework on a real humanoid robot and demonstrate its effectiveness, flexibility, and safety in two different scenarios and more than 20 tasks.

\section{Problem Formulation}
\label{sec:formulation}


In this work, we consider the interaction as a leader-follower formulation~\cite{vianelloHumanHumanoidInteractionCooperation2021}, where the human is the leader and the humanoid robot is the follower.
Define $\mathcal{I}$ as the set of human intentions and $\mathcal{K}$ as the set of robot skills.
At time step $t$, the leader shows an intention $I_t \in \mathcal{I}$ for the follower to perform a skill $K_t \in \mathcal{K}$, such as picking up a can, brushing a plate, or stamping a file.
We assume one intention corresponds to at most one skill, and the robot should be able to switch between skills in real time. The map function is defined as $f: \mathcal{I} \rightarrow \mathcal{K}$.

The skills of the robot can be categorized into three types: interactive motion, manipulation, and idle.
The interactive motion skills require the robot to perform expressive and diverse behavior, and the manipulation skills require the robot to interact with objects in the environment precisely. 
When the human leader does not show any intention, the robot will be in an idle state and do nothing.
The real-time interaction design requires the robot to respond to the leader's intentions with low latency and in-skill reflection to the human leader.
Low latency requires the robot to predict the leader's intention in real time and interrupt the current skill when the leader shows a new intention.
In-skill reflection requires the robot to react to human motion and environment even if the human intention is not changed, such as pausing current movement if the robot collides with the human or the target object is not reachable.



We formulate the observation space $\mathcal{O}$ of a humanoid robot as the combination of the
environment state $\mathcal{E}$ and the human behavior $\mathcal{H}$, i.e., $\mathcal{O} = \mathcal{E}  \bigodot  \mathcal{H} $.
The environment state $\mathcal{E}$ includes the robot's proprioception and the object state. The human behavior $\mathcal{H}$ consists of the leader's intention and the leader's behavior. 
To reduce the complexity of observation, our framework decomposes interaction policy into several sub-modules and separately designs the observation space for each module.
Compared to end-to-end models, this decomposed design allows the robot to learn humanoid-human-object interaction from human-object-human demonstration data, which is more sample-efficient and scalable.

\section{\our Framework}
\begin{figure*}[ht]
    \centering
    \includegraphics[width=\linewidth]{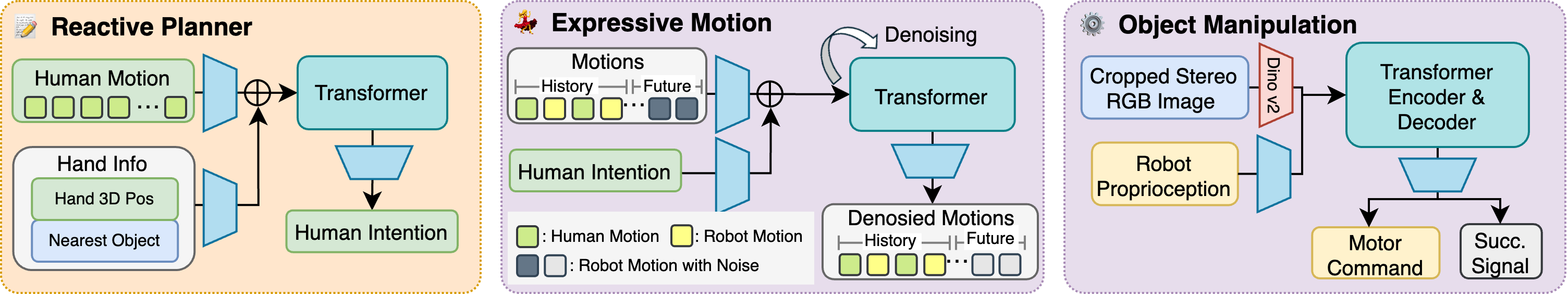}
    \caption{\textbf{Network architecture of \our modules}, including the reactive planner, motion generation, and manipulation skills.}
    \label{fig:method}
\end{figure*}

In \our, the humanoid robot acts as the follower and learns to predict the human intention $I$ with the reactive planner, then utilizes the corresponding skill $K$ to finish the interaction.
Those skills are classified into interactive motion, manipulation, and idle. 
Interactive motion skills, simply called motion skills, enable the robot to react to the leader's intentions with real-time motion. 
Manipulation skills enable the robot to handle objects based on the predicted intentions. 
Idle refers to the robot maintaining its joints in a default state.
\Cref{fig:formulation} illustrates the framework of \our, and \Cref{fig:method} provides the detailed network architecture of each sub-module in our implementation.


\

\subsection{Data Collection}
\noindent\textbf{Human-object-human interaction data.}
To learn the interaction between humans and robots, we first collect a dataset of human-object-human interaction~\cite{zhang2024core4d}, where two people perform a series of daily interaction tasks with various objects.
In comparison to human-robot interaction data, human-object-human interaction data can be collected without a real robot, which is cheaper to collect and easier to scale to more skills in various of scenarios.
The dataset contains interaction between two people in two different scenarios, dining and office. 
The dataset is recorded with a simple motion capture system, and a stereo RGB-D camera in the first-person view of the follower. 
The motion capture system that collects the follower's behavior is described in \Cref{app:mocap}.
Motion data is retargeted to the humanoid robot and used by imitation learning algorithms to construct the reactive motion skills.
The stereo RGB-D camera records the leader's behavior $\mathcal{H}$ and the environment state $\mathcal{E}$, which is used to predict the leader's intention $I$.

We label each frame $t$ in the interaction data with the leader's intention $I_t$ and the follower's skill $K^{(t)}$, which are represented as ID integers of intentions and skills. We add additional labels for the occupancy $p \in \mathcal{P}$ of the robot, 
indicating whether the end-effector (i.e., the hands) is empty or interacting with an object, with distinct labels assigned to different objects.

\noindent\textbf{Teleoperation data.}
Different from the interactive motion skills, 
certain skills, such as picking up a cup, require more precise control of the robot's end-effector and the manipulation of some objects.
To ensure the success of those skills, 
we collect demonstrations with a teleoperation system~\cite{cheng2024tv}, where the human's motion is captured with a VR device, and the robot's joint positions are set by retargeting the human's motion.
This system records the control commands, the robot's proprioception, and stereo videos from a camera on the robot's head to perceive the environment.
We also label the frames where the skill showcase is completed successfully, referred to as the success signal, which is used to learn the finish condition of the manipulation skills.

\begin{figure*}[ht]
    \centering
    \includegraphics[width=\linewidth]{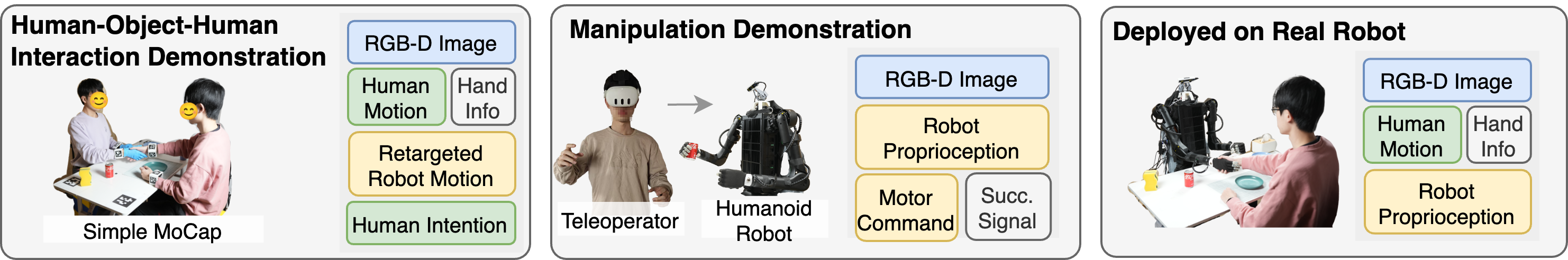}
    \caption{\textbf{Information collected in the dataset and the requirement of real robot deployment.} The left box shows our Human-Object-Human Interaction Demonstration Data, which contains human motion and hand position information collected from RGB-D images and a simple motion capture system. 
    The box in the middle illustrates our Manipulation Demonstration Data, which is collected through teleoperation. 
    The right box shows the information required in real robot deployment.}
    \label{fig:data}
\end{figure*}

\Cref{fig:data} shows information collected in the interaction and teleoperation dataset and the requirement of real robot deployment. More details are described in \Cref{app:implementation}.

\subsection{Reactive Planner}
\label{sec:reactive-planner}
The reactive planner is designed to infer the leader's intention $I_t$ from the real-time observation $O_t$ and decide the next skill $K^{(t+1)}$ of the robot.
The planner is a Transformer model, which takes the human's motion and the environment information as input, and predicts the leader's intention at a 30Hz frequency.
To enhance the generalization of the model, we do not input observed images directly, but extract the human's body and hand postures, the human's hand and head position, and the nearest object to hands from the RGB-D images, as well as the robot's hand occupancy $p_t$.
We retarget human hand postures to a robot hand with 6 degrees of freedom (DoF) and represent the hand posture as the position of each joint.

There are two types of skills that correspond to the leader's intentions: motion skills and manipulation skills. Each skill $K$ has a start condition $s_K \in \mathcal{P}$ and an end transition $e_K \in \mathcal{P}$. 
The start condition shows the required hand occupancy of each hand to start the skill. For example, the skill to cheer with the leader requires the humanoid to hold a can of drink in the right hand.
The end transition determines the change of hand occupancy after finishing this skill successfully. For example, for the skill of picking up a can, the start condition and end transition are \texttt{[empty, empty]} and \texttt{[empty, can]} respectively.
A comprehensive description of all skills is shown in \Cref{app:skills}.

The humanoid robot starts from an idle state. If the reactive planner predicts a human intention $I$ consistently for $n_{r}$ time steps, the humanoid robot switches to the corresponding skill $T=f(I)$.
The human intentions are meant to ``start'' the execution of a skill, rather than ``keep'' the current execution. For example, the leader only needs to point at the can for a while at the beginning to get the robot to pick the can on the table, instead of pointing all the time. 

After a skill is initiated, the motion skill persists until a change in human intention occurs, while the manipulation skill persists until the execution is judged successful or exceeds a time limit.
When a skill is complete, the robot returns to the idle state.

To enable low-latency interaction, the application of most of the skills can be interrupted by another skill when a different human intention lasts $k_2$ steps. 
The motion skills can be easily undone by immediately moving to an idle pose, while the interruption of the manipulation skills is complicated as it requires reversing the object state.
We use a corresponding reverse skill to interrupt each interruptible manipulation skill. For example, the skill of placing the can is a reversal of the skill of picking the can.
We show the detailed transitions between skills in \Cref{app:implementation}.

When the current occupancy is not satisfied with the start condition of a skill $p_t \neq s_K$, the skill is not able to start.
To satisfy the requirement, we 
We build a directed graph of occupancy transition. The node $n\in \mathcal{P}$ is hand occupancy and the edge $e\in \mathcal{T}$ is skills. 
Before starting to demonstrate a skill $K$ with an unsatisfied condition, we find the shortest path from current occupancy $p_t$ to the start condition $s_K$, and execute the skill series $\{K_1, K_2, \ldots  \}$ in order.
After the operations mentioned above are done, the target skill $K$ can be utilized. The occupancy graphs of 2 scenarios are shown in~\Cref{app:implementation}.

\subsection{Interactive Motion Skills}
\label{sec:inter-motion-skill}
In humanoid-human interactions that do not involve complex object manipulation, the primary objective of the humanoid robot is to produce smooth, consistent motions while providing robust real-time feedback on human behavior. 
To accomplish this, we employ a multi-body motion diffusion model~\citep{liang2024intergen} to generate low-level interactive motion skills. 

Different from multi-person motion generation, the humanoid and human are heterogeneous and asymmetric in the humanoid-human interaction.
We represent the human motion $m^1_t$ as a 6D rotation vector for each joint, and the humanoid motion $m^2_t$ as the target of humanoid robot joint positions. Both motions are simplified to arm and hand joints. We also add hand occupancy $p_t$ and human intention $I_t$ as input to the model, to ensure the robot's motion is consistent with the human's intention. Details can be found in \Cref{app:implementation}.

Our model predicts the future motion of the humanoid robot $m^2_{t+1:t+5}$ based on the history of human motion $m^1_{t-30:t}$ and humanoid motion $m^2_{t-30:t}$.
The model predicts 5 future frames of humanoid motion with a 3 Hz frequency, which generates 30 frames of motion in one second.

The network structure is a Transformer-based model, which takes the loss of reconstructing the humanoid motion as the main loss, and the velocity loss of motion as an auxiliary loss.

\subsection{Manipulation Skills}
\noindent\textbf{Imitation learning.}
To enhance the smoothness and robustness of interactive motion reactions, the motion generation model described in Section~\ref{sec:inter-motion-skill} omits RGB images as input. However, this design makes the model insufficient for dexterous object manipulation, resulting in a lower success rate in practice. 
Meanwhile, as mentioned earlier, the retargeting inevitably introduces deviations between the humanoid’s end-effector poses and original human motions, further leading to manipulation failures.

As a result, we train independent Action Chunking Transformer (ACT)~\citep{zhao2023learning} models for each low-level manipulation skill. 
The ACT model enables 30Hz real-time inference of the robot's joint positions.
Demonstrations collected by teleoperation are manually segmented and labeled as distinct skills for model training.
To satisfy low-latency requirements, we trained a paired reverse skill model for each manipulation skill, enabling the robot to handle human interruptions properly.

\noindent\textbf{Learning terminal conditions.}
In our multi-skill interactive manipulation framework, the model must recognize when a current skill is completed in order to transition to the next skill. 
In addition to desired humanoid joint positions, each manipulation policy predicts an additional success signal. The signal is an indicator of whether the skill is completed, and we add an extra cross-entropy loss to train the 0/1 classification.

\noindent\textbf{Robust and safe manipulation.}
We crop the image to the region of robot-object interaction as the manipulation model input. The cropped image input removes the leader human's body and only keeps the hand information, which helps the model focus on the manipulation skill and be robust to human appearance and behavior changes.
For skills with a single arm, the input/output of the model only includes the corresponding arm information.
We also collect in-skill interruption data, where the robot pauses or withdraws its current movement if it collides with the human or the target object is unreachable.
Such data helps the robot to exhibit safe behavior and in-skill reflection to the change of human behavior or environment, even if the human intention is not changed.

\subsection{Safety Supervisor}

A safety supervisor serves as a global guarantee of safe robot actions, which forces the robot to pause immediately when potential harm is detected.
In this module, the collision box of the robot is calculated based on several selected key points on the arms.
They are updated by forward kinematics as the movement of the arm.
Meanwhile, global coordinates of human hands are obtained by the depth camera.

The safety module judges whether the robot collision box is to collide with human hands. 
In case the distance between them is too close, the safe supervisor sends an unsafe signal to halt the robot at the current pose until the distance recovers into a safe range.
We use the Euclidean distance from human hand key points to robot arm key points as a simple but effective approach to calculating the collision box. 

The safe supervisor takes strong aids to ensure the robot would not hurt humans by avoiding collision, especially in skills where they should contact at a rather close distance such as handshake and handover the plate.

\subsection{Real Robot Platform}
\begin{figure}[ht]
    \centering
    \includegraphics[width=0.7\linewidth]{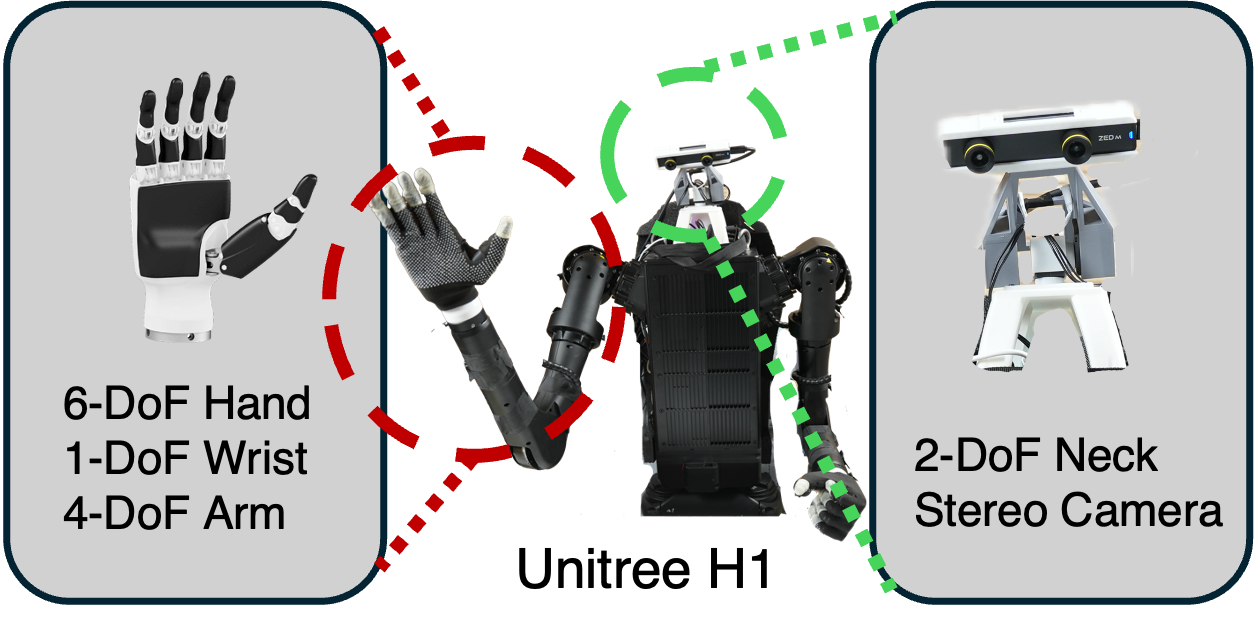}
    \caption{The humanoid robot platform.}
    \label{fig:humanoid}
\end{figure}

\label{method:real-robot}

We implement \our on a Unitree H1 humanoid robot with an active head, equipped with a RGB-D stereo camera ZED mini, illustrated in~\Cref{fig:humanoid}.
The human body posture is detected from a body detection model in ZED API with 38 joints and represented as 6D rotation vectors of joints. 
We capture the leader human's intention and the state of objects from the RGB-D images.
The hand posture is detected by a hand reconstruction model, HaMeR~\citep{pavlakos2024reconstructing}.
We record the information of nearest objects to hands, including the object ID, distance to the hand, and the IOU of the object bounding box and the hand bounding box.
The 3D bounding box of objects and hands is detected from a stereo RGB-D camera, ZED-mini, with a fine-tuned YoloV11~\citep{khanam2024yolov11} model and a body detection model in ZED API. 
The RGB-D stereo camera enables retrieval of the 3D position of any 2D pixel in the image, represented in world coordinates. The orientation of the humanoid's body determines the X and Y axes, while gravity defines the Z axis.


\begin{table*}
    \begin{minipage}[t]{0.48\linewidth}
        \begin{center}
            \caption{\textbf{Performance of the interaction planner.}}
            \label{tab:planner}
            \resizebox{\linewidth}{!}{
            \begin{tabular}{c|cc|cc|c}
                \toprule
                \multirow{3}{*}{\textbf{Method}} & \multicolumn{4}{c|}{\textbf{mAP} $\uparrow$} & \multirow{3}{*}{\shortstack{\textbf{Inference} \\ \textbf{Frequency} $\uparrow$}}\\
                \cmidrule{2-5}
                & \multicolumn{2}{c}{\textbf{Validation Data}} & \multicolumn{2}{|c|}{\textbf{Test Data}} & \\    
                & \textbf{Scene 1} & \textbf{Scene 2} & \textbf{Scene 1} & \textbf{Scene 2} & \\
                \midrule
                Ours & 0.982 & \textbf{1.0} & \textbf{0.787} & \textbf{0.643} & \textbf{30 Hz} \\
                Ours (w/o hand details) & \textbf{0.999} & 0.925 & 0.729 & 0.587 & \textbf{30 Hz} \\
                Qwen2-VL-2B-In. & - & - & 0.213 & 0.167 & \textbf{30 Hz} \\
                Finetuned Qwen2-VL-2B-In. & 0.284 & 0.322 & 0.228 & 0.159 & \textbf{30 Hz} \\
                ChatGPT-4o-mini & - & - & 0.573 &  0.564 & $\approx 0.46$ Hz  \\
                \bottomrule
            \end{tabular}}
        \end{center}
    \end{minipage}%
    ~~
    \begin{minipage}[t]{0.52\linewidth}
        \begin{center}
            \caption{\textbf{Performance of motion generation.}}
            \label{tab:motion}
            \resizebox{\linewidth}{!}{
            \begin{tabular}{c|cccc}
            \toprule
            \textbf{Method} & \textbf{FID$\downarrow$} & \textbf{JPE(mm)$\downarrow$} & \textbf{Diversity}&\textbf{MModality$\uparrow$}\\
                \midrule
                Real &  - & - & 3.74 \textpm 0.05 & -\\
                \midrule
                Zero Velocity & 43.22 \textpm 0.01 & 84.82 \textpm 0.01 & 2.85 \textpm 0.09 & - \\
                \midrule
                Ours & \textbf{10.67 \textpm 0.01} & \textbf{48.79 \textpm 0.00} & \textbf{3.68 \textpm 0.06} & 0.02 \textpm 0.00 \\
                Ours (w/o diffusion) & 38.50 \textpm 0.01 & 142.85 \textpm 0.02 & 2.91 \textpm 0.04 & - \\
                Ours (w/o human motion) & 17.34 \textpm 0.05 & 60.52 \textpm 0.04 & 3.59 \textpm 0.08 & \textbf{0.06 \textpm 0.01} \\
                \bottomrule
            \end{tabular}}
        \end{center}
    \end{minipage}
\end{table*}



\section{Experiment}
In experiments, we evaluate the performance of \our on two different scenarios: a dining waiter scenario and an office assistant scenario, also called Scene 1 and Scene 2. The details of the skills in each scenario are shown in ~\Cref{app:skills}. The robot should react with its arms, hands, and active head. 
We first evaluate the performance of each module in \our, including human intention prediction, motion generation, and manipulation skills. 
To show the effectiveness of our framework, we compare it with end-to-end models on different numbers of skills. 
The results show that all modules in \our perform well in the skills. The framework outperforms the end-to-end models and is more robust to out-of-distribution data.
We also analyze the failure of the system and show the crucial challenges in the real-world deployment of humanoid robots.

\subsection{Framework Performance}
\noindent\textbf{1) Human Intention Prediction}

We evaluate the performance of our human intention prediction module by calculating the mAP score on datasets. 
We split the training data into 80\% for training and 20\% for validation, and also collected a test set under the robot-human-interaction deployment setting.
The test set contains 3 different people, each of whom performs all intentions of 2 scenarios, which ensures the diversity of the test set.
\Cref{tab:planner} shows that although deploying the model to the real world 
leads to a decrease in performance, but our model still outperforms all baselines.

Our method takes both \textit{human motion} and \textit{hand details} including hand 3D position and object nearest to hand as input. Compared to the baseline that only inputs human motion, our method performs better in our test set, demonstrating that \textit{hand details} is necessary information to differentiate between different human intentions. We also test the performance of VLMs on human intention prediction. Qwen2-VL-2B-Instruct can infer at a frequency of 30Hz, but its performance is poor even after being finetuned on our training set, probably due to a relatively small amount of training data. While GPT-4o-mini can perform quite well on our test set, it takes too long for an inference which leads to a slow reaction of the robot. 
When calculating mAP for the VLMs, we assume that the probability of the class output by VLM is 1, while the probabilities of all other classes are 0.

\noindent\textbf{2) Motion Generation}

We compare our motion generation module with three baselines on all the motions involved in our skills (\textit{handshake}, \textit{wave}, \textit{cheers}, \textit{thumbup}, \textit{spread hand}, \textit{take photo}). The baselines are:
\begin{itemize}[leftmargin=*]
    \item \textbf{Zero Velocity}: the repetition of the last pose observed, as a simplest baseline.
    \item \textbf{Ours (w/o diffusion)}: Generate the motions directly with a Transformer model with the same structure as the denoiser used in our Diffusion-based method, without the full diffusion framework.
    \item \textbf{Ours (w/o human motion)}: Generate humanoid motions only conditioning on the human intention label and history of humanoid motions, without the guidance of detailed human motion.
\end{itemize}

The metrics to evaluate the performance of the motion generation modules are:
\begin{itemize}[leftmargin=*]
    \item \textbf{FID}: 
    The FID score~\cite{heusel2017gans} is leveraged to assess the similarity between synthesized and real motions quantitatively.
    \item \textbf{JPE}: 
    We calculate the Joint Position Error (JPE) based on the forward kinematic results of the generated robot joint 3D position to measure the poses of all the individuals. JPE is averaged over all hand joints and finger joints.
    \item \textbf{Diversity}: 
    We calculate the average Euclidean distances of $300$ randomly sampled pairs of motions in latent space to measure motion diversity in the generated motion dataset. The Diversity of motions generated by the model is expected to be closer to the Diversity of Real Data.
    \item \textbf{MModality}: 
    MModality captures the ability of the model to generate diverse motions for the same human intention label and human motion sample.
    We sample 20 motions within one fixed human intention label and one fixed sample of human motion to form 10 pairs, and measure the average latent Euclidean distances of the pairs. The average overall human intention and human motion pairs are reported. 
\end{itemize}

~\Cref{tab:motion} shows the performance of the motion generation module, where \our significantly outperforms all baselines on FID and JPE, demonstrating better quality of generated motions. The baseline without the diffusion process gets the lowest score, indicating that the sampling process of the diffusion model helps to synthesize better motions.
Diversity also matters since different people may react differently to the same motion, and we expect our robot to obtain diverse behavior as well. As is shown in \Cref{tab:motion}, our approach outperforms the baseline without diffusion in terms of Diversity, thanks to the stochasticity introduced by the diffusion process and the ability of diffusion models to fit high-dimensional distributions. The baseline model without human motion inputs gets a higher MModality score, because it may generate quite different motions given only a human intention label. Yet it performs poorer on FID
and JPE, since human intention alone can not guide the model to generate expressive motions with accurate and desired reactive meaning. For instance, the policy may give various angles and poses of a stretched hand given only the intention \texttt{Shake Hands}; in contrast, given the exact human pose and hand positions, our model can stretch the hand to the exact position and shake hand with the human, highlighting its real-time reactive capability.


\begin{table*}
\begin{minipage}[t]{0.56\linewidth}
    \begin{center}
        \caption{\textbf{Performance of manipulation across objects.}
        }
        \label{tab:manipulation}
        \resizebox{\linewidth}{!}{
        \begin{tabular}{c|cccc|cccc}
        \toprule
        \multirow{2}{*}{\textbf{Metrics}} & \multicolumn{4}{c|}{\textbf{Scene 1}} & \multicolumn{4}{c}{\textbf{Scene 2}} \\
        & \textbf{\textit{Can}} & \textbf{\textit{Plate}} & \textbf{\textit{Sponge}} & \textbf{Tissue} & \textbf{\textit{Cap}} & \textbf{\textit{Book}} & \textbf{\textit{Stamp}} & \textbf{\textit{Lamp}}\\ 
            \midrule
Success Rate & 1.00 & 0.96 & 0.90 & 0.95 & 0.93 & 0.95 & 0.93 & 1.00 \\
Average Time & 9.41 & 29.59 & 23.69 & 9.43 & 16.14 & 10.81 & 15.17 & 5.06 \\
            \midrule 
Success Rate (Human) & 0.97 & 0.98 & 0.99 & 0.91 & 0.91 & 0.93 & 0.92 & 0.96 \\
Average Time (Human) & 10.42 & 25.77 & 17.04 & 9.54 & 18.98 & 10.21 & 11.84 & 3.53 \\
            \bottomrule
        \end{tabular}}
    \end{center}
    \end{minipage}
    ~~~~
    \begin{minipage}[t]{0.4\linewidth}
    \begin{center}
        \caption{\textbf{Success rate of manipulation} with different ratios of interrupted data.}
        \label{tab:safety_manipulation}
        \resizebox{\linewidth}{!}{
        \begin{tabular}{c|ccc}
        \toprule
        \textbf{Ratio of data} & \multirow{2}{*}{\textbf{Pick Can}} &  \multirow{2}{*}{\textbf{Stamp the Paper}} & {\textbf{Place Plate}} \\
        \textbf{with interruption} & & & \textbf{to Stack} \\
            \midrule
1$\%$ & 0.00 & 0.00 & 0.00   \\
10$\%$ & 0.05 & 0.15 & 0.30  \\
20$\%$ & 0.85 & 0.60 & 0.90   \\
            \bottomrule
        \end{tabular}}
    \end{center}
    \end{minipage}
\end{table*}


\noindent\textbf{3) Objects Manipulation}
\label{sec:exp-manip}

We further test the manipulation performance, and collect the results in \Cref{tab:manipulation}, comparing with statistics recorded from human teleoperation. We compute the success rate and the averaged time based on $20$ independent tests, on the main object in two different scenes.
The detailed statistics of each certain skill can be found in \Cref{app:result}.

Our manipulation module shows good performance aligned with human teleoperators. 
As only success data of human teleoperation is used in training, the module even outperforms humans by success rate on groups with rather simple motions such as \textit{can}, \textit{tissue}, \textit{book} and \textit{lamp}. 
In skills requiring a more delicate operation, the trained model slightly falls behind. 
We mainly consequence the fault for the heterogeneity between humans and the robot. 
The inadequate DoFs of the robot arms and lack of haptic sensing on the dexterous hand add great difficulty to some of the skills.
To be specific, the former would lead to the failure of a smooth trajectory, ending with the cap sticking on the hatstand (in group \textit{cap}), and the latter makes it challenging for the robot to determine whether the stamp is pressed to a fair location (in group \textit{stamp}).

It is also noteworthy that the average operation time of our manipulation module is slightly longer on most of the skills than that of human data. 
This is because (1) some skills have a periodical motion, making the progress predictor output ending progress value later than the ground truth. (2) the robot would slowly move back to a fair initialized joint position range at the start to avoid violent actions, this adds to lags as a trade-off.

To determine the effectiveness of the in-skill interruption data, we analyze the impact of training data with human disturbance on three typical manipulation motions. 
With a fixed amount of total training data, 1$\%$, 10$\%$, or 20$\%$ of them are replaced by data where the robot motion is disturbed.
Take the skill \textit{Pick Can} as an example: the robot finds a human is looting the can when it intends to pick, then the robot should withdraw its motion and try to pick again when the human returns the can.

The success rate of a proper motion is shown in \Cref{tab:safety_manipulation}.
The result shows that with an increasing ratio of disturbance data, the success rate of dealing with human violations is getting higher. 
Models with few disturbance data are not capable of a withdraw action, and only mixing as a ratio of 20$\%$ could make a success rate of 85$\%$ in simple skills.

\noindent\textbf{4) Framework Structure}

\begin{figure}[t]
    \centering
    \includegraphics[width=\linewidth]{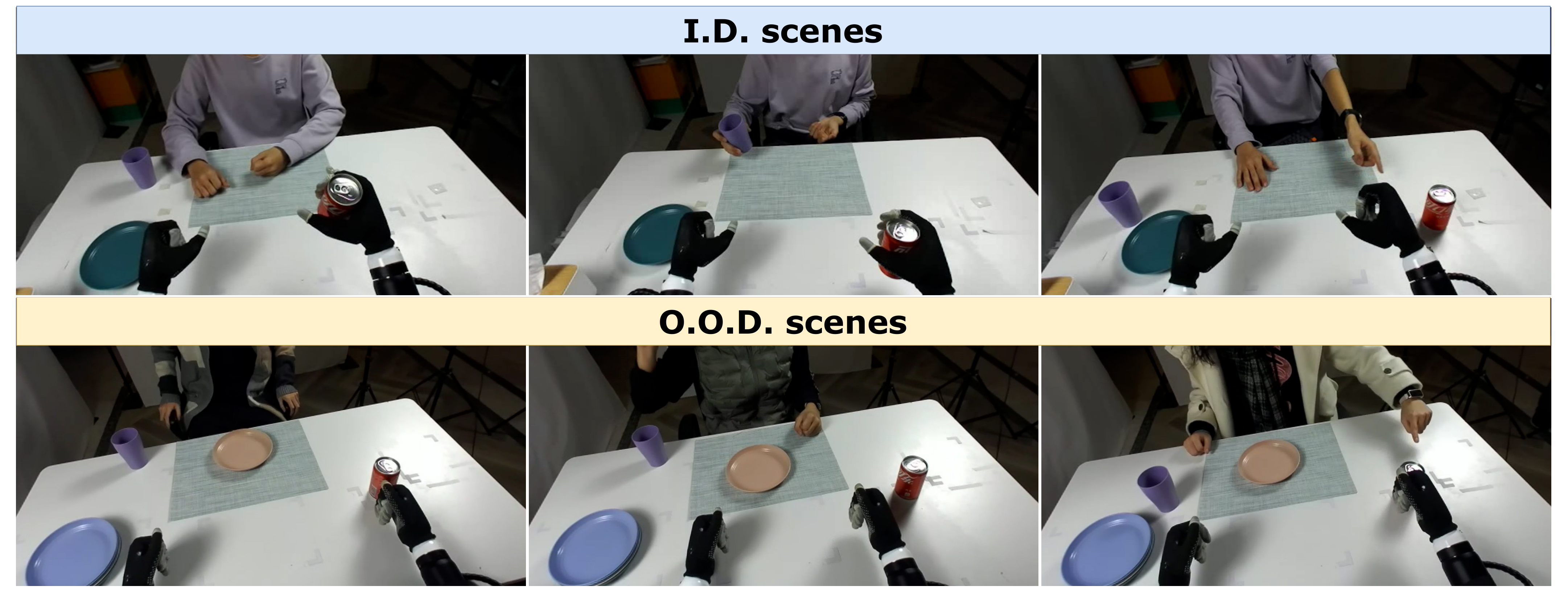}
    \caption{
    \textbf{Examples of I.D. and O.O.D. scenes comparing \our with the End-to-End Model from the robot camera view.}
    The first row (I.D. scenes) shows the leader human wearing consistent clothing, with only minor adjustments in the arrangement of objects on the table.
    The second row (O.O.D. scenes) presents leader humans in different clothing and entirely different object arrangements, representing diverse scenarios encountered during deployment.
    }
    \label{fig:e2e_scene}
\end{figure}

\begin{table}
    \begin{center}
        \caption{\textbf{Success rates of \our and end-to-end model} with a different number of skills.
        }
        \label{tab:framework}
        \begin{tabular}{c|ccc}
        \toprule
        \textbf{Number of tasks} & $\mathbf{1}$ \textbf{skill} & $\mathbf{3}$ \textbf{skills} & $\mathbf{5}$ \textbf{skills} \\
            \midrule
Ours & \textbf{1.00} & \textbf{0.95} & \textbf{0.97} \\
End2End (I.D. Scene) & \textbf{1.00} & 0.70 & 0.84 \\
End2End (O.O.D. Scene) & 0.95 & 0.57 & 0.67 \\
            \bottomrule
        \end{tabular}
        \vspace{-10pt}
    \end{center}
\end{table}
To highlight the accessibility to multiple skills of our \textbf{two-level} framework, we also compare end-to-end (E2E) baselines, similar to the setting of \citet{cardenas2024xbg}. 
We implement an ACT policy~\cite{zhao2023learning,cheng2024tv}, leaving the input image not cropped to capture the human intention and robot motion information correctly.
As a concise case, the baselines are trained on data of only $1$ motions (\textit{cheers}), $3$ skills (along with \textit{pick}, \textit{place}, meaning pick or place the can) and $5$ skills (along with \textit{handshake}, \textit{wave} also) in a simple scene, and are named E2E/1, E2E/3 and E2E/5 separately. 
In dataset collection, $100$ slices for each skill are collected separately, the same as an average number of that in the manipulation skills.
In deployment, we test the E2E model on the in-distribution (I.D.) scene in which human clothing and object arrangement are the same as the recorded datasets, and in the out-of-distribution (O.O.D.) scene these conditions vary.

In detail, the I.D. scene features a leader human wearing a light purple shirt, with a Coke can placed to the right of the robot, and a dark green plate on the table.
The O.O.D. scenes include, but are not limited to, the following changes: replacing the Coke can (red) with a Sprite can (green), swapping the dark green plate for a light red or light blue one, adding more plates in front of the leader human, changing the leader human's outfit, for example, to a down jacket, or replacing the leader human with another character, such as one in a striped sweater or a white coat.
\Cref{fig:e2e_scene} shows an example highlighting the scene difference with respect to the leader human.

The success rates on average of different numbers of skills are shown in \Cref{tab:framework}, where we can observe \our outperforms the E2E baselines on the skills of better prediction of human intention and robustness to O.O.D. data. 
When trained with data of only one skill, the E2E framework is able to predict human intention and perform a successful motion. 
However, when more different skills are used, the E2E framework struggles to predict the correct intention.
Meanwhile, without the information on hand occupancy, the model is likely to fail to tell the difference among skills with similar camera view, i.e. \textit{cheers}, \textit{pick}, and \textit{place}.
Moreover, the coupling of prediction, generation, and manipulation leads to uncropped images as input.
The high dimension of images and noise cripple the robustness of the prediction module.
Interacting with O.O.D. human body and clothing, or under some O.O.D. table arrangement, the model fails to generate proper motion to finish the skill showcase.
The detailed success rate is shown in \Cref{app:result}.
.
\subsection{Analysis of System Failure}
As a framework of multiple modules, the failure of the system could be caused by various reasons. 

\noindent\textbf{Error and limitation of sensors.} 
Most of the perception of our implementation of \our is based on one RGB-D camera. However, the estimation of 3D position often shifts with time and missing when the estimated object is occluded by other objects or the robot arms. The cumulative error of the sensors leads to a misunderstanding of human intention and incorrect judgment of the safety supervisor.

\noindent\textbf{Stability of hardware.}
The zero position of the robot arm may have shifted in a small range, which leads to the incorrect proprioception.
Also, the robot's electronics age over time, which causes errors that require precise control of the robot's end-effector.

\noindent\textbf{Failure of Model Generalization.}
Due to the limited data collection, the model may fail to generalize to extreme out-of-distribution scenarios, although we mitigate this issue by cropping the image to the region of interest for manipulation skills and using extracted information rather than raw images for human intention prediction.
Some unseen human clothing or unexpected object arrangement may lead to the failure of the manipulation.
Also, non-standard sitting posture or body shape can also have an impact on the prediction of human posture and intention, which leads to misunderstanding of human intention.
Fortunately, human leaders can intervene to correct the robot's behavior in \our, which helps prevent a complete breakdown of the system.
\section{Conclusion and Limitations}

In this work, we presented \our, a hierarchical learning framework designed to enable humanoid robots to engage in real-time humanoid-human-object interactions. By decoupling the interaction process into high-level planning and low-level reactive control, \our allows humanoid robots to quickly adapt to the change of human intentions and interrupt ongoing tasks without delays. This framework incorporates a wide variety of skills, from object manipulation to expressive motion generation. We implemented \our on a real humanoid robot and demonstrated its effectiveness, flexibility, and safety in various dynamic environments.

The proposed framework provides a significant step toward making humanoid robots autonomous and responsive in real-world applications, such as assisting with daily life tasks, disaster response, and industrial automation. By allowing continuous humanoid-human-object interaction, \our provide immediate and adaptive responses, making humanoid robots suited for seamless integration into human environments.

Despite promising results, several limitations remain. First, while \our is designed to be scalable, the current implementation is constrained by the availability of high-quality training data. The generalization of the system across a broader range of tasks and environments is still a challenge, as it heavily relies on human demonstrations and teleoperation data, which is time-consuming to collect. Future work will focus on utilizing existing datasets and simulation environments to improve the scalability and generalization of the framework.
Additionally, the current implementation of \our is limited to the upper body at a fixed workspace, but a humanoid assistant should have locomotion and navigation abilities in a dynamic environment, and react with whole-body behaviors. Future work should integrate a whole-body controller to extend the framework to whole-body interaction for humanoid robots, and more general tasks with varying levels of human intervention.





{
    \small
    \bibliographystyle{plainnat}
    \bibliography{references}
}

\clearpage
\appendix
\label{sec:appendix}

\subsection{Real World Setup}

\noindent\textbf{Deployment hardware}
The humanoid robot on which we deploy \our is Unitree H1~\cite{H1-page}.
Following \citet{cheng2024tv}, we assembled two DYNAMIXEL XL330-M288-T motors~\citep{dynamixel-page} with 3D printed gimble parts and a ZED Mini stereo camera ~\cite{zed-page} for two-DoF (yaw and pitch) active sensing.
Each arm of H1 has 5 DoFs and a 6-DoF end-effector from~\cite{dexterous-page}, and other DoFs on the robot are not used.

\begin{figure}[b]
    \centering
    \includegraphics[width=\linewidth]{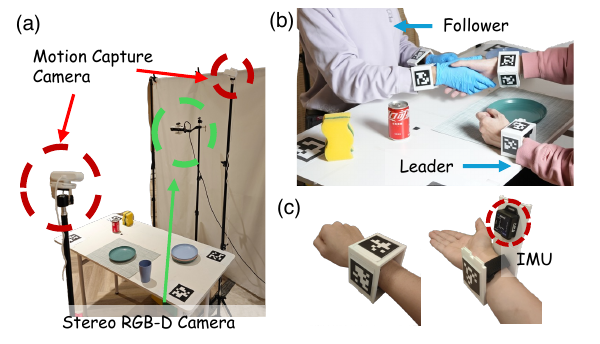}
    \caption{\small \textbf{Setup of motion capture system.} a) The two Motion Capture Cameras are used to detect the ArUco markers. The video recorded by the RGB-D Camera is used to process human motion and human hand details of the Leader. b) Follower and Leader both wear wristbands with ArUco markers for hand position detection. c) We 3D-print our wristbands with 4 ArUco markers on 4 surfaces and embed a IMU beneath the upper surface.}
    \label{fig:mocap_sys}
\end{figure}

\noindent\textbf{Motion capture system}
\label{app:mocap}
We use ArUco markers and two cameras to build a simple motion capture system.
We put four ArUco markers on the four corners of the workspace table to locate the cameras.
Each human in the workspace wears two 3D-printed wristbands with four ArUco markers on each of them, illustrated in \Cref{fig:mocap_sys}. 
The cameras are calibrated and located using the OpenCV library and capture the human wristband's position in real time.
Each wristband has an additional IMU sensor to capture the orientation of the human's wrist.
To reduce the noise in the collected data, we use a Kalman filter to smooth the data.

\begin{figure}
    \centering
    \includegraphics[width=0.85\linewidth]{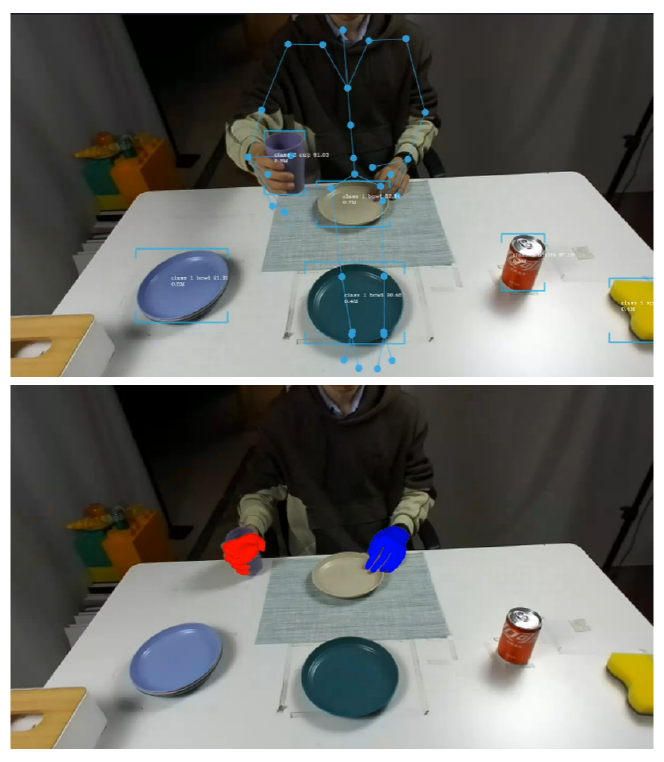}
    \caption{\small \textbf{Visualization results of human body motion detection, object detection, and human hand detection.} The upper image demonstrates the human motion detection (only upper body is used) and object detection (in Dining scenario). The lower image demonstrate the human hand detection.}
    \label{fig:zed_detect}
\end{figure}

\noindent\textbf{Motion Detection and Object Detection}
As is illustrated in \Cref{method:real-robot}, we use the Body Tracking feature in ZED API to detect the body motion of the human and a fine-tuned YoloV11~\cite{khanam2024yolov11} model to detect objects on the table. For hand detection, we use HaMeR\cite{pavlakos2024reconstructing} to obtain the human hand motion and then retarget\cite{qin2024anyteleopgeneralvisionbaseddexterous} it to the 6-DoF robot hand. The visualization results are shown in \Cref{fig:zed_detect}.

\subsection{Skill Descriptions}
\label{app:skills}
In this section, we describe the skills that we deploy on the humanoid robot. The details include the description, success condition, reverse skill (if exists), and the human intention related to the skill.
Note that the intention is inferred mainly from human behavior, hand positions, and the relative location of objects.
The latter two can be concluded trivially to the start condition and end transition, and are shown by skill in \Cref{tab:skills}.
We concern the intention mainly on the human body motion in the narration as follows:

\noindent\textbf{1) Scenario 1: Humanoid as a Dining Waiter}

In this scenario, the leader human and the robot sit face-to-face at the side of a dining table. 
There are plates with food, a Coke can, a tissue box, and a sponge on the table.
In the following skill descriptions, the humanoid robot takes the role of a helpful waiter and serves the leader human a meal with these objects.
$10$ skills related to $4$ objects are listed below:

\begin{itemize}[leftmargin=*]
    \item \textit{Pick Can}: The robot picks up a can with its right arm from the table. 
    The success condition is that the can is lifted off the table.
    The interruption data includes the human taking the can away or the human putting his hand on the can. 
    \textit{Place Can} is the corresponding reverse skill.
    The leader human shows the intention by pointing to the can when the right robot arm is empty.
    
    \item \textit{Place Can}: The robot places the can back on the table with its right arm. 
    The success condition is that the can is placed on the table and the robot's hand is lifted off the can.
    The intention of this skill is shown by the leader human pointing to the place on the table where the can was placed before.
    
    \item \textit{Get Plate from Human}: The robot fetches a plate from the hand of the human with its left arm.
    The success condition is the plate in the dexterous hand when the human loosens the grip of the plate.
    The leader human shows the intention by handing a plate forward.
    
    \item \textit{Place Plate to Stack}: The robot stacks the plate in its right hand onto a pile of plates on the table.
    The success condition is the plate settled on the top of the plate pile without slipping.
    The intention is given by the leader human pointing to the stack.
    Interruption data, in which the human touches the plate to stick the motion, is added to the collected dataset.
    
    \item \textit{Pick Place from Table}: The robot lifts a plate on the table by both arms and holds it in the left hand.
    Application of this skill succeeds if there is no slippage in the motion until the plate is held.
    The leader human points to the plate to show the intention.
    
    \item \textit{Handover Plate}: The robot protracts its left arm to give the leader human the plate on it.
    The showcase of this skill ends when the plate is put into the human hand.
    The leader human simply stretches the right hand out to show the intention.
    It is the reverse skill of \textit{Get Plate from Human}.
    
    \item \textit{Pick Sponge}: The robot picks up the sponge with its right arm. 
    The sponge is placed beside the can.
    When it is lifted off the table the skill showcase succeeds.
    The leader human shows intention by mimicking washing.
    Data where the human snatches the sponge before the robot reaches it adds to the dataset. 
    In case this happens during deployment, the robot withdraws its hand to the idle state.

    \item \textit{Brush with Sponge}: It is a complex skill using both arms.
    The start condition is a plate in the left hand and a sponge in the right one when the leader human makes the washing gesture (same as that in \textit{Pick Sponge}) again.
    To apply this skill, the robot moves the sponge close to the plate and rubs the sponge on the plate to brush it.
    The success condition is the robot keeps the periodic brushing motion for over $10$ seconds.

    \item \textit{Place Sponge}: The reverse skill of \textit{Pick Sponge}.
    The robot puts the sponge in the right hand back onto the table to complete the skill demonstration.
    The intention is shown by the leader human pointing to the place on the table where the sponge was placed before (similar to skill \textit{Place Can}).

    \item \textit{Pick a Piece of Tissue}: The leader human points to the tissue box to express the intention. 
    Then the robot uses its left hand to pull a tissue from the tissue box placed on the table corner and gives it to the leader human.
    The skill showcase succeeds when the leader human receives the tissue.
\end{itemize}

\begin{table*}
    \begin{center}
        \caption{Description of the skills. Notes: The \textbf{Start Condition} or \textbf{End Transition} \texttt{[A, B]} means that object \texttt{A} is in the left hand of the robot and \texttt{B} is in the right hand. \texttt{empty} for this hand must be empty, \texttt{any} for this hand could hold any object or be empty, and \texttt{-} for this object remains unchanged after the skill is completed.}
        \label{tab:skills}
        \begin{tabular}{ccccccc}
        \toprule
        \textbf{Scenarios} & \textbf{Object} & \textbf{Skill Name} & \textbf{Start Condition} & \textbf{End Transition} & \textbf{Num. of Data} & \textbf{Arm}\\
            \midrule
            \multirow{11}{*}{\begin{tabular}{c}
                 Scenario 1 \\
                 Dining Waiter
            \end{tabular}}  & \multirow{2}{*}{can} & Pick Can & \texttt{[any, empty]} & \texttt{[-, can]} & $107$ & Right\\
            & & Place Can & \texttt{[any, can]} & \texttt{[-, empty]} & $100$ & Right\\
            \cmidrule{2-7}
            & \multirow{4}{*}{plate} & Get Plate from Human  & \texttt{[empty, any]} & \texttt{[plate, -]} & $100$ & Left\\
            & & Place Plate to Stack & \texttt{[plate, any]} & \texttt{[empty, -]} &  $98$ & Left\\
            & & Pick Plate from Table & \texttt{[empty, empty]} & \texttt{[empty, plate]} & $115$ & Dual-Arm\\
            & & Handover Plate & \texttt{[plate, any]} & \texttt{[empty, -]} & $115$ & Left\\
            \cmidrule{2-7}
            & \multirow{3}{*}{sponge} & Pick Sponge  & \texttt{[any, empty]} & \texttt{[-, sponge]} & $89$ & Right\\
            & & Brush with Sponge & \texttt{[plate, sponge]} & \texttt{[-, -]} & $81$ & Dual-Arm\\
            & & Place Sponge & \texttt{[any, sponge]} & \texttt{[-, empty]} & $82$ & Right\\
            \cmidrule{2-7}
            & tissue & Pick a Piece of Tissue  & \texttt{[empty, any]} & \texttt{[-, -]} & $105$ & Left\\
            \midrule
        \multirow{8}{*}{\begin{tabular}{c}
             Scenario 2  \\
             Office Assistant
        \end{tabular}} & \multirow{2}{*}{cap} & Settle Cap & \texttt{[any, empty]} & \texttt{[-, -]} & $111$ & Right\\
            & & Handover Cap & \texttt{[any, empty]} & \texttt{[-, -]} & $110$ & Right\\
            \cmidrule{2-7}
            & book & Pick Book & \texttt{[empty, any]} & \texttt{[-, -]} & $115$ & Left\\
            \cmidrule{2-7}
            & \multirow{3}{*}{stamp} & Pick Stamp & \texttt{[any, empty]} & \texttt{[-, stamp]} & $92$ & Right\\
            & & Stamp the Paper & \texttt{[any, stamp]} & \texttt{[-, -]} & $87$ & Right\\
            & & Place Stamp & \texttt{[any, stamp]} & \texttt{[-, empty]} & $89$ & Right\\
            \cmidrule{2-7}
            & lamp & Turn off/on the Lamp & \texttt{[empty, any]} & \texttt{[-, -]} & $85$ & Left\\
            \midrule 
            \multirow{6}{*}{\begin{tabular}{c}
                 Expressive Motions
            \end{tabular}
            } & \multirow{6}{*}{None} & Cheers & \texttt{[any, can]} & \texttt{[-, -]} & $66$ & Dual-Arm\\
            & & Wave & \texttt{[any, empty]} & \texttt{[-, -]} & $39$ & Dual-Arm\\
            & & Shake Hands & \texttt{[any, empty]} & \texttt{[-, -]} & $51$ & Dual-Arm\\
            & & Take Photo & \texttt{[any, empty]} & \texttt{[-, -]}  & $31$ & Dual-Arm\\
            & & Thumb Up & \texttt{[empty, empty]} & \texttt{[-, -]}  & $22$ & Dual-Arm\\
            & & Spread out Hands & \texttt{[empty, empty]} & \texttt{[-, -]} & $26$ & Dual-Arm\\
            \bottomrule
        \end{tabular}
    \end{center}
\end{table*}

\noindent\textbf{2) Scenario 2: Humanoid as an Office Assistant}

In this scenario, the leader human sits across the humanoid robot at an office table.
This time the robot transforms into an office assistant and deals with complicated cases such as stamping paper for approval, settling a baseball cap on the rack, picking and handing a book over, and reacting properly if the human takes a snap in working.
There are $7$ skills related to $4$ objects in this scenario.

\begin{itemize}[leftmargin=*]

\item \textit{Settle Cap}: The robot gets a cap from the leader human's hand and settles it on a hat rack with its right arm.
The skill showcase begins with the human holding the cap with both hands and ends with the robot pulling its hand back from the hat rack.

\item \textit{Handover Cap}: The robot takes the cap off the hat rack and sends it to the leader human.
It is the reverse skill of \textit{Settle Cap}.
The related intention is inferred when seeing the human pointing to the rack.
The success condition is that the human has received the cap.

\item \textit{Pick Book}: The robot picks a book from the shelf and hands it over.
The skill begins with the human gesturing toward the book.
When the human takes the book, this skill is completed successfully.

\item \textit{Pick Stamp}: The robot picks up the stamp on the table with its right hand. 
The skill succeeds when the stamp is lifted near the hand in an idle posture.
The leader human instructs the execution of this skill by passing along the paper.

\item \textit{Stamp the Paper}: It is a delicate operation to make an issue for approval.
The robot presses the stamp down onto the paper to mark a sign.
This skill is considered successful only if one mark is imprinted.
It is noteworthy that printing more than one mark in a single execution means that the model fails to predict the ending, thus being treated as a failure case.
The sign of the related intention is the leader human pointing at the paper.
To make an in-skill interruption, the human covers the paper with a hand to make the robot withdraw its hand if the pressing is not done.

\item \textit{Place Stamp}: The robot places the stamp back with its right hand.
It is the reverse skill of \textit{Pick Stamp} and is triggered by withdrawing the paper.

\item \textit{Turn off/on the Lamp}: Turning on and off the lamp share the same motion, and thus are trained as one skill.
When the human slumps over the office desk to take a nap, the robot taps the switch of the lamp to turn off it.
And when the human wakes up and lifts the head, the robot operates the same motion to turn on the lamp.
    
\end{itemize}

\noindent\textbf{3) Interactive Motion Skills}

Some skills are not involved with object operation and, thus, are not trained as manipulation skills. 
They are noted as motion skills.
These skills are considered successful when the robot performs the motion properly as the human shows the intention and recovers to the idle posture when the intention no longer sustains.

\begin{itemize}[leftmargin=*]
\item \textit{Cheers}: The robot reaches out the right hand to touch the bottle held by the right hand of the human.
Though holding a Coke can in the right hand during deployment, the robot does not manipulate the object.
For this reason, this skill is not trained in a manipulation demonstration model.

\item \textit{Wave}: The robot lifts up the right hand and waves the right hand when the leader human is waving also.

\item \textit{Shake Hands}: The robot stretches its right hand out to touch the hand of the leader human with a handshaking posture.

\item \textit{Take Photo}: The robot lifts up the right hand and makes a V-sign when the human raises the phone to take a photo, and puts the hand done as the human puts away the phone.

\item \textit{Thumb Up}: The robot reaches both hands out with the thumbs up as the human gives it a thumb-up.
Human intention with the left hand, right hand, or both is approved.

\item \textit{Spread out Hands}: The robot stretches its arms out to the sides with palms up when the leader human spreads its hands out.
    
\end{itemize}

\subsection{Prompt for VLMs}
Here are the prompts we give to Qwen and GPT-4o-mini in evaluation of the intention prediction module.

\noindent\begin{tcolorbox}[
    colframe=darkgray, 
    boxrule=0.5pt, 
    colback=lightgray!20, %
    arc=3pt, 
    fontupper=\small,
    breakable, title={Prompt for the \textbf{Dining} scenario},
    ]
    
\textit{You are a humanoid robot sitting in front of a human and equipped with a camera slightly tilted downward on your head, providing a first-person perspective. I am assigning you a new task to recognize to human gestures in front of you. Remember, the person is sitting facing you, so be mindful of their gestures. If the person is holding a cup to you and trying to cheer with you, answer `Cheers'. If the person is giving you a thumbs-up, answer `Thumbup'. If the person extends their right hand to shake hands with you, answer `ShakingHand'.If the person is waving to you with the right hand, answer `Waving`. If the person is taking a photo of you with a cellphone, answer `Taking Photo`. If the person is spreading out both hands in a gesture of resignation, answer `Spreading Hands`. If the person is pointing to a Coke can in the middle of the table (on your right side), answer `Pointing Can'. If the person is pointing to an empty spot on the table with no objects (on your right side), answer `Pointing Table`. If the person is pointing to a tissue box at the far left of the table, answer `Pointing TissueBox'. If the person is pointing to a plate in the middle of the table (just in front of you), answer `Pointing Plate'. If the person is holding out the right hand with the palm open toward you, answer `Palmup'. If the person is handing you a plate, answer `Handing Plate'. If the person is clenching their right fist, holding their left hand open and upward, and placing their right hand above the left as if pretending to wash a plate, answer `Washing'. If the person is pointing at a stack of plates on the left side of the table, answer `Pointing Plates'. If the person is pointing at a sponge on the right side of the table, answer `Pointing Sponge'. If the person is crossing his arms to form an X shape, answer `Cancel'. If no significant gestures are made, answer `Idle'. 
Respond directly with the corresponding options [Cheers, Thumbup, ShakingHand, Pointing Can, Pointing TissueBox, Pointing Plate, Palmup, Handing Plate, Washing, Pointing Plates, Pointing Sponge, Cancel, Idle] based on the current image and observed gestures. Directly reply with the chosen answer only, without any additional characters.}
\end{tcolorbox}

\noindent\begin{tcolorbox}[
    colframe=darkgray, 
    boxrule=0.5pt, 
    colback=lightgray!20, %
    arc=3pt, 
    fontupper=\small,
    breakable, title={Prompt for the \textbf{Office} scenario},
    ]
\textit{You are a humanoid robot sitting in front of a human and equipped with a camera slightly tilted downward on your head, providing a first-person perspective. I am assigning you a new task to recognize human gestures in front of you. Remember, the person is sitting facing you, so be mindful of their gestures. If the person is giving you a thumbs-up, answer `Thumbup`. If the person extends their right hand to shake hands with you, answer `ShakingHand`. If the person is waving to you with the right hand, answer `Waving`. If the person is taking a photo of you with a cellphone, answer `Taking Photo`. If the person is spreading out both hands in a gesture of resignation, answer `Spreading Hands`. If the person is handing you a cap, answer `Handing Cap`. If the person is pointing at a cap place on the right of the table, answer `Pointing Cap`. If the person is handing a document to you with both hands and you are NOT holding a stamp, answer `Handing File`. If a document is placed in the center of the table in front of you, and the person is pointing to it with the right hand, answer `Pointing File`. If the person retrieves the document from your side of the table to the other side, directly across from you, and you are still holding the stamp, answer `Retrieve File`. If the person is lying down on the table and the lamp is ON, answer `Lie Down`. If the person is sitting up from the table and the lamp is OFF, answer `Sit up`. If the person is pointing at the books standing in the top left corner of the table, answer `Pointing Book`. If the person is crossing the arms to form an X shape, answer `Cancel`. If no significant gestures are made, answer `Idle`. You are NOT holding a stamp right now and the lamp is now ON, observe the image and gestures carefully. Respond directly with the corresponding options [Pointing Book, Handing Cap, Pointing Cap, Handing File, Pointing File, Retrieve File, Lie Down, Sit up, ShakingHand, Thumbup, Cancel, Idle]. Directly reply with the chosen answer ONLY, without any additional characters.}
\end{tcolorbox}

The sentence `\textit{You are NOT holding a stamp right
now and the lamp is now ON}' is modified at each query according to the current situation (whether the robot is holding a stamp and whether the lamp is on).

\subsection{Implementation of RHINO Modules}

\label{app:implementation}

\noindent\textbf{1) Reactive Planner}

\begin{algorithm}
\caption{Pseudo-code for Skill Transitions of Reactive Planner.}
\begin{algorithmic}[1]
\label{alg:planner}
\STATE $Skill \gets \text{Idle}$
\WHILE{$true$}
    \STATE $human\_intention \gets \text{Recognize\_Human\_Intention()}$
    \IF{human intention is stable for $k$ frames and human intention != current intention}
        \IF{human intention $=$ Idle and Skill $=$ Manipulation}
            \STATE Continue
        \ENDIF
        \IF{Skill $=$ Manipulation and interruptionAllowed}
            \STATE $Skill \gets \text{Reverse\_Skill}(Skill)$
        \ENDIF
        \IF{$\text{Start\_Condition}(human\_intention)$ is not satisfied by hand occupancy}
             \STATE $path \gets$  FindPath(occupancy, StartCondition $(human\_intention))$
            \STATE $Skill \gets \text{Execute\_Path}(path)$
        \ELSE
            \STATE $Skill \gets \text{Corresponding Skill} (human\_intention)$
        \ENDIF
    \ELSIF{SkillSucceeded(Skill) or SkillTimeout(Skill)}
        \STATE $Skill \gets \text{Idle}$
        \IF{SkillSucceeded(Skill)}
            \STATE $\text{Hand Occupancy} \gets \text{End\_Transition}(Skill)$
        \ENDIF
    \ENDIF
\ENDWHILE
\end{algorithmic}
\end{algorithm}

    
\begin{figure}[t!]
    \centering
    \begin{subfigure}[t]{\linewidth}
        \centering
        \includegraphics[width=\linewidth]{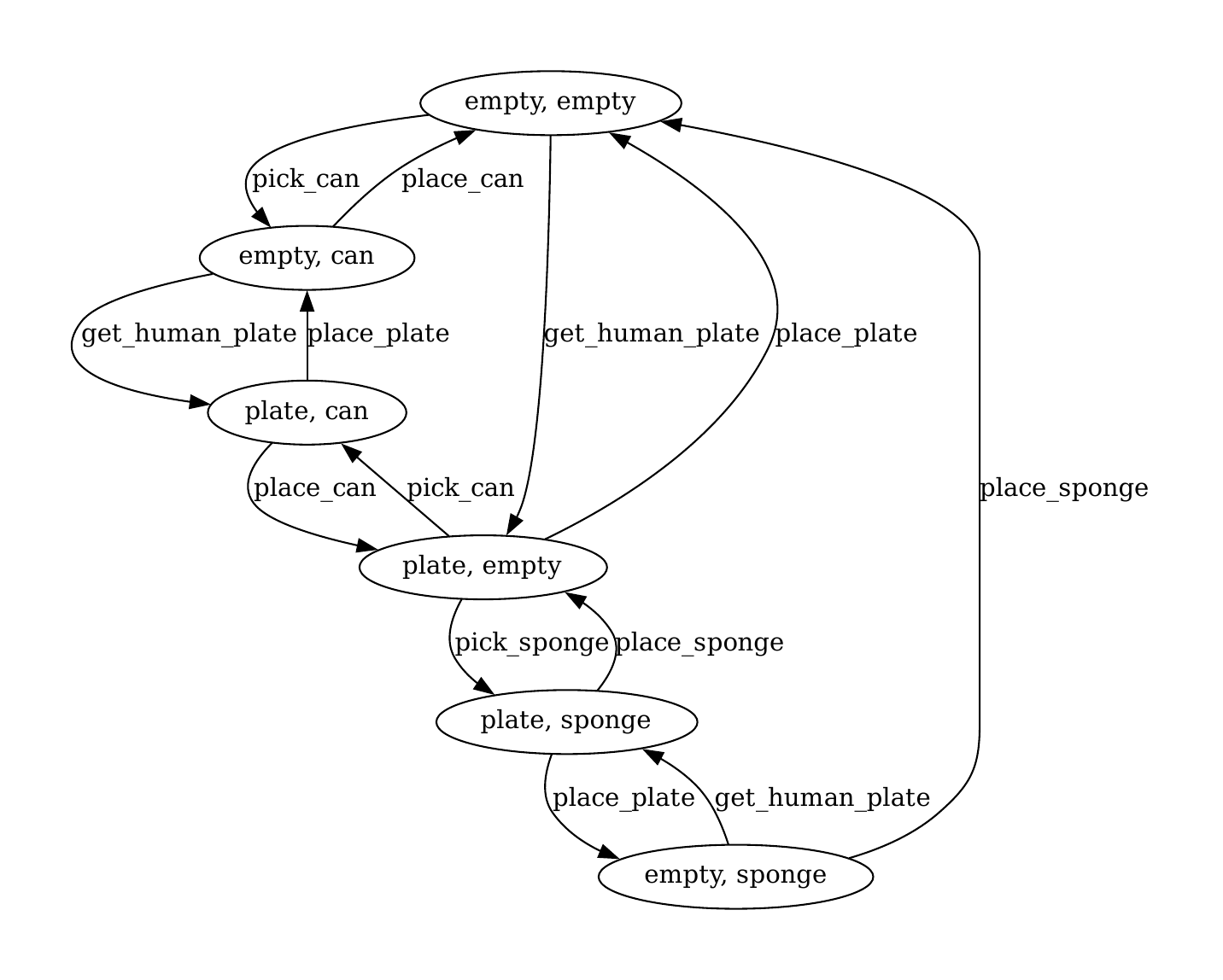}
        \caption{Dining scenario.}
    \end{subfigure}%
    \\
    \begin{subfigure}[t]{\linewidth}
        \centering
        \includegraphics[width=0.8\linewidth]{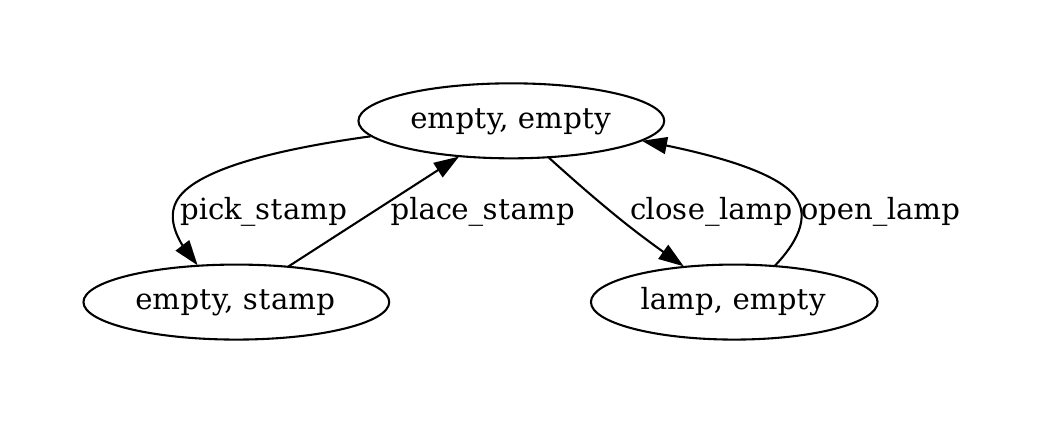}
        \caption{Office scenario.}
    \end{subfigure}
    \caption{Occupancy graph of two different scenarios.}\label{fig:occupancy}
\end{figure}

The switching logic of the skill planner is listed in ~\Cref{alg:planner}.
The directed graphs of occupancy are shown in~\Cref{fig:occupancy}.
Here we further explain the \texttt{Recognize\_Human\_Intention()} function in detail, which is implemented as a transformer-based classifier. The model input includes:
\begin{itemize}[leftmargin=*]
    \item \textbf{Upper Body Human Posture}: a 36-dim human upper body skeleton, namely the 6D rotation of the wrist, elbow and shoulder joints for each arm.
    \item \textbf{Human Hand Pose}: a 12-dim human hand pose vector. For each hand, we retarget the detected human hand pose to our robot hand with IK, and take the 6 joint pos as human hand pose vector.
    \item \textbf{Robot Hand Occupancy}: a 10-dim robot hand occupancy label. Since we have at most 5 objects in total (Can, Cup, Plate, Sponge, Tissue), we use a 5-dim one-hot label for each hand to represent the object held in the robot's hand. If the robot is not holding anything, the label will be all-zeros.
    \item \textbf{Human Details}: a 19-dim vector, including the x and y-axis of each human hand position, the z-axis (height) of the human head position, and a 7-dim label for the nearest object to each hand. The nearest object label is concatenated by a 5-dim one-hot label of the object type, the distance from the object to the human hand, and the average of IOU and IOFs of the object bounding box and the human hand bounding box.
\end{itemize}
We use an MLP encoder to encode the concatenated vector of  \textbf{Upper Body Human Posture}, \textbf{Human Hand Pose} and \textbf{Robot Hand Occupancy}, and another MLP to encode \textbf{Human Details} to latent dimension. The concatenated latent vector is processed by a Transformer backbone, followed by a final MLP layer to predict the human intention class. The 
hyper-parameters of the Transformer backbone are listed in ~\Cref{tab:planner-hyper}.


\begin{table*}[t]
    \begin{minipage}[t]{0.3\linewidth}
        \begin{center}
            \caption{Hyper-parameters of the Reactive Planner.}
            \label{tab:planner-hyper}
            \resizebox{0.9\linewidth}{!}{
            \begin{tabular}{cc}
                \toprule
                hyper-parameter&value\\
                \midrule
                latent dimension&128\\
                num head&8\\
                num layers&3\\
                batch size&256\\
                feed-forward dimension&128\\
                maximum epoch&300\\
                learning rate&0.0001\\
                \bottomrule
            \end{tabular}
            }
        \end{center}
    \end{minipage}
    \hfill
    \begin{minipage}[t]{0.3\linewidth}
        \begin{center}
            \caption{Hyper-parameters of the Motion Generation Model.}
            \label{tab:motion-hyper}
            \resizebox{0.9\linewidth}{!}{
            \begin{tabular}{cc}
                \toprule
                hyper-parameter&value\\
                \midrule
                latent dimension&256\\
                num head&8\\
                num layers&4\\
                feed-forward dimension&256\\
                diffusion steps&300\\
                sampling steps&30\\
                batch size&512\\
                maximum epoch&4000\\
                learning rate&0.0001\\
                \bottomrule
            \end{tabular}
            }
        \end{center}
    \end{minipage}
    \hfill
    \begin{minipage}[t]{0.3\linewidth}
        \begin{center}
            \caption{Hyper-parameters of the ACT model for manipulation skills.}
            \label{tab:act-hyper}
            \resizebox{0.9\linewidth}{!}{
            \begin{tabular}{cc}
                \toprule
                hyper-parameter&value\\
                \midrule
                KL weight&10\\
                Cross-entropy weight&1\\
                chunk size&30\\
                hidden dimension&512\\
                batch size&45\\
                feed-forward dimension&3200\\
                maximum epoch&50000\\
                learning rate&0.00005\\
                \bottomrule
            \end{tabular}
            }
        \end{center}
    \end{minipage}
\end{table*}

\noindent\textbf{2) Interactive Motion Generation}

For interactive motion generation, we use a transformer-based diffusion model, which denoises the past 30 frames of human and robot motions and future 5 frames of robot motions. Both human motion and robot motion consist of upper-body motion (36-dim for humans and 10-dim for humanoid), hand motion (6-dim for each hand,) and hand occupancy label (5-dim one-hot label for each hand). Besides, the predicted human intention label is also conditioned during the diffusion process. The hyper-parameters of our model are listed in \Cref{tab:motion-hyper}.

\noindent\textbf{3) Manipulation Skills}

Thanks to the stability of model training, most of the hyper-parameters are basically consistent across all skills.
The volume of data for training each skill is shown as a column in \Cref{tab:skills}.
The hyper-parameters in training ACT models~\cite{zhao2023learning} are shown as \Cref{tab:act-hyper}.

\begin{table*}[htb]
    \begin{center}
        \caption{Performance of manipulation module across manipulation skills. }
        \label{tab:manipulation_detailed}
        \begin{tabular}{ccccccc}
        \toprule
        \textbf{Scenarios} & \textbf{Object} & \textbf{Skill Name} & \textbf{Success Rate} & \textbf{Average Time} & \begin{tabular}{c}
             \textbf{Success Rate}  \\
             \textbf{(Human)}
        \end{tabular} & \begin{tabular}{c}
             \textbf{Average Time}  \\
             \textbf{(Human)}
        \end{tabular} \\
            \midrule
            \multirow{11}{*}{\begin{tabular}{c}
                 Scenario 1 \\
                 Dining Waiter
            \end{tabular}}  & \multirow{2}{*}{can} & Pick Can & 1.00 & 5.31 & 1.00 & 5.77\\
            & & Place Can & 1.00 & 4.10 & 0.93 & 4.65\\
            \cmidrule{2-7}
            & \multirow{4}{*}{plate} & Get Plate from Human  & 1.00 & 4.86 & 0.98 & 5.12 \\
            & & Place Plate to Stack & 0.95 & 8.19 & 0.97 & 6.91 \\
            & & Pick Plate from Table & 0.90 & 10.75 & 0.96 & 8.60 \\
            & & Handover Plate & 1.00 & 5.79 & 1.00 & 5.14 \\
            \cmidrule{2-7}
            & \multirow{3}{*}{sponge} & Pick Sponge  & 0.95 & 8.19 & 1.00 & 7.45 \\
            & & Brush with Sponge & 0.90 & 10.02 & 1.00 & 4.18 \\
            & & Place Sponge & 0.85 & 5.57 & 0.98 & 5.41 \\
            \cmidrule{2-7}
            & tissue & Pick a Piece of Tissue  & 0.95 & 9.43 & 0.91 & 9.54\\
            \midrule
            \multirow{8}{*}{\begin{tabular}{c}
             Scenario 2  \\
             Office Assistant
            \end{tabular}} & \multirow{2}{*}{cap} & Settle Cap & 1.00 & 7.50 & 0.91 & 8.50\\
            & & Handover Cap & 0.85 & 8.64 & 0.90 & 10.48 \\
            \cmidrule{2-7}
            & book & Pick Book & 0.95 & 10.81 & 0.93 & 10.21 \\
            \cmidrule{2-7}
            & \multirow{3}{*}{stamp} & Pick Stamp & 1.00 & 4.80 & 0.92 & 3.91 \\
            & & Stamp the Paper & 0.80 & 5.64 & 0.92 & 3.11 \\
            & & Place Stamp & 1.00 & 4.74 & 0.93 & 4.83 \\
            \cmidrule{2-7}
            & lamp & Turn off/on the Lamp & 1.00 & 5.06 & 0.96 & 3.53 \\
            \bottomrule
        \end{tabular}
    \end{center}
\end{table*}

For the prediction of the success signal, we marked the last $n_{s}$ frames of the recorded data as $1$ (completed) and other frames as $0$ to generate a 0/1 label.
$n_{s}$ is set to $25$ in most of the skills and shifted to $10$ in three of them of which the ending frames changed sharply in motion.
The special skills are \textit{Pick Stamp}, \textit{Stamp the Paper}, and \textit{Place Stamp}.

\noindent\textbf{4) Safety Supervisor}

\begin{figure}[t]
    \centering
    \includegraphics[width=0.9\linewidth]{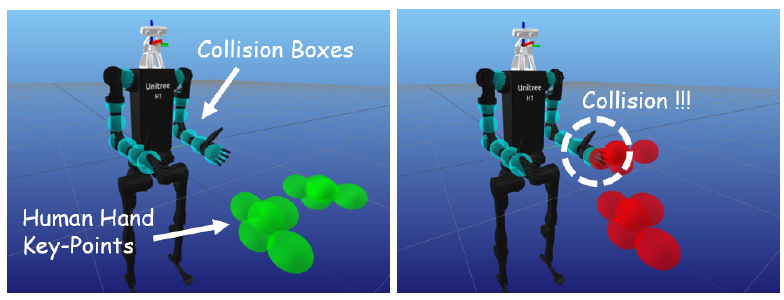}
    \caption{The interface displays the work of the safety supervisor, with sphere markers representing the collision boxes of the human hands and robot arms. These markers move in sync with the interaction. When an unsafe collision is detected, the human hand markers change color from green to red.}
    \label{fig:safe}
\end{figure}

The collision box is calculated using $14$ key points across each arm. The key points at specific joints and their midpoints are identified as follows:

\begin{itemize}[leftmargin=*]
    \item The origins of the shoulder pitch, shoulder yaw, elbow, and wrist joints are defined as key points.
    \item Additional key points include the midpoints between the shoulder yaw and elbow joints, and between the elbow and wrist joints.
    \item A further key point is defined at one-third the distance beyond the elbow towards the wrist, extending from the segment between these two joints.
\end{itemize}
This structured delineation allows for precise calculations pertinent to robotic arm movements within a predefined spatial configuration.

The human hands are shaped by the detected key points from body detection model of ZED API, from which each hand is reconstructed as $5$ points.
Once one of the points is close to any robot key point in $0.1$ meters, an unsafe signal is broadcast to pause the robot control.

We also provide the visualization of the safety supervisor, of which the interface shown in \Cref{fig:safe}. When the human hand key-points collide with any collision box, the supervisor will send an unsafe signal to halt the robot.
Our safety supervisor runs at $30$Hz.

\subsection{Detailed Experiment Results}

\label{app:result}

\noindent\textbf{1) Planner}
\begin{figure*}[t]
    \vspace{-3.0em}

    \centering
    \begin{minipage}{0.48\textwidth}
        \centering
        \includegraphics[height=8.5cm, keepaspectratio]{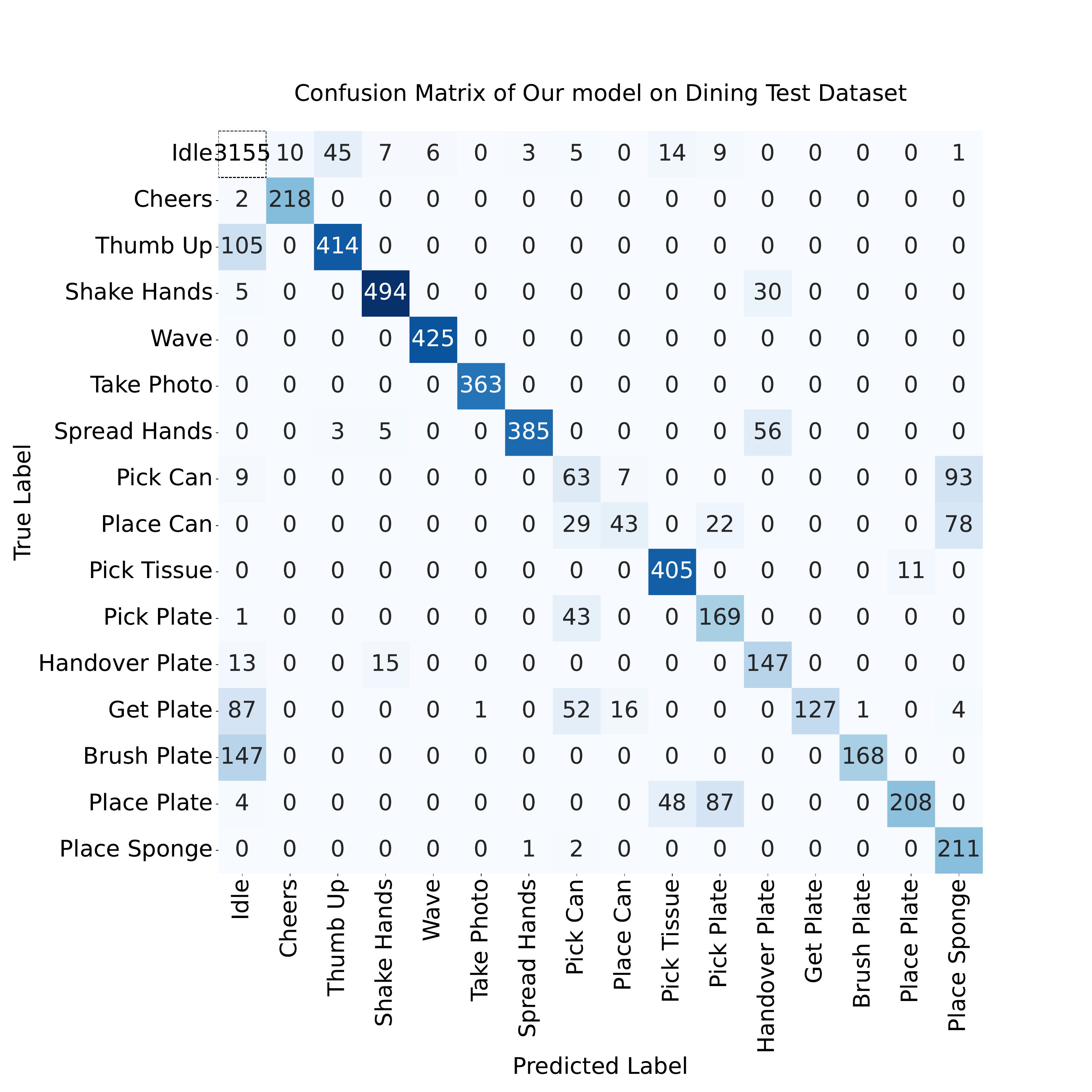}
        \label{fig:image1}
    \end{minipage}
    \begin{minipage}{0.48\linewidth}
        \centering
        \includegraphics[height=8.5cm, keepaspectratio]{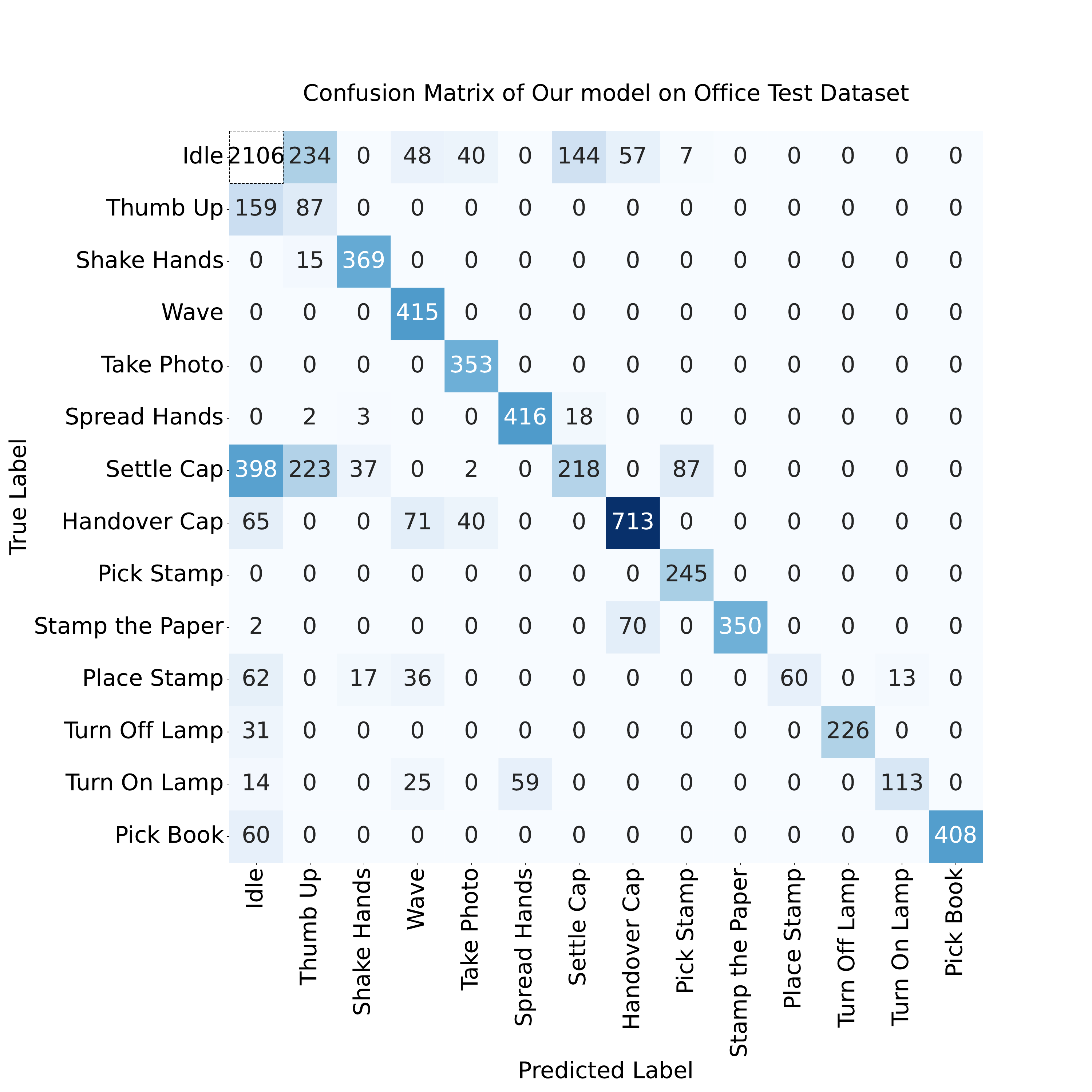}
        \label{fig:image2}
    \end{minipage}

    \vspace{-1.5em}

    \begin{minipage}{0.48\textwidth}
        \centering
        \includegraphics[height=8.5cm, keepaspectratio]{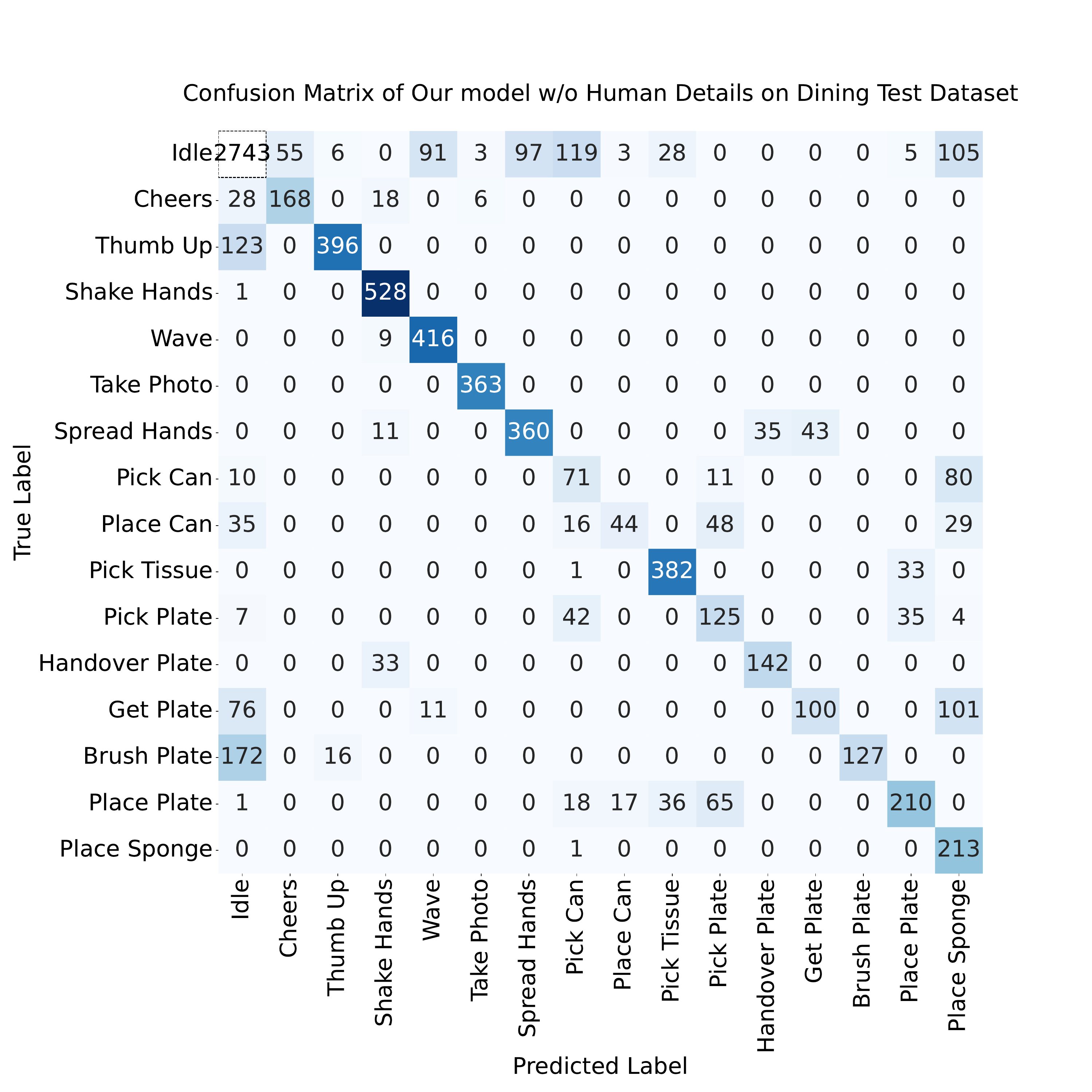}
        \label{fig:image3}
    \end{minipage}
    \begin{minipage}{0.48\linewidth}
        \centering
        \includegraphics[height=8.5cm, keepaspectratio]{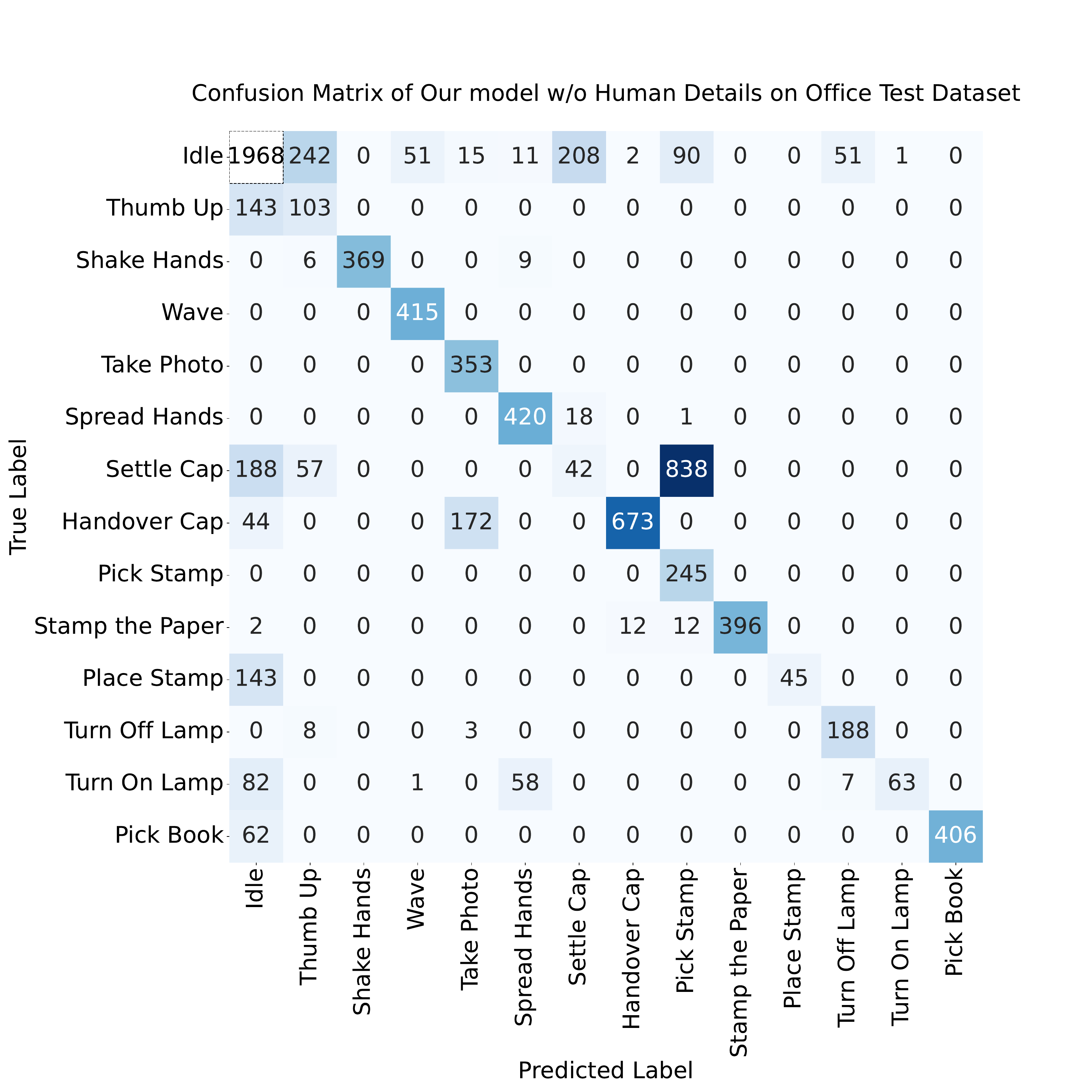}
        \label{fig:image4}
    \end{minipage}

    \vspace{-1.5em}

    \begin{minipage}{0.48\textwidth}
        \centering
        \includegraphics[height=8.5cm, keepaspectratio]{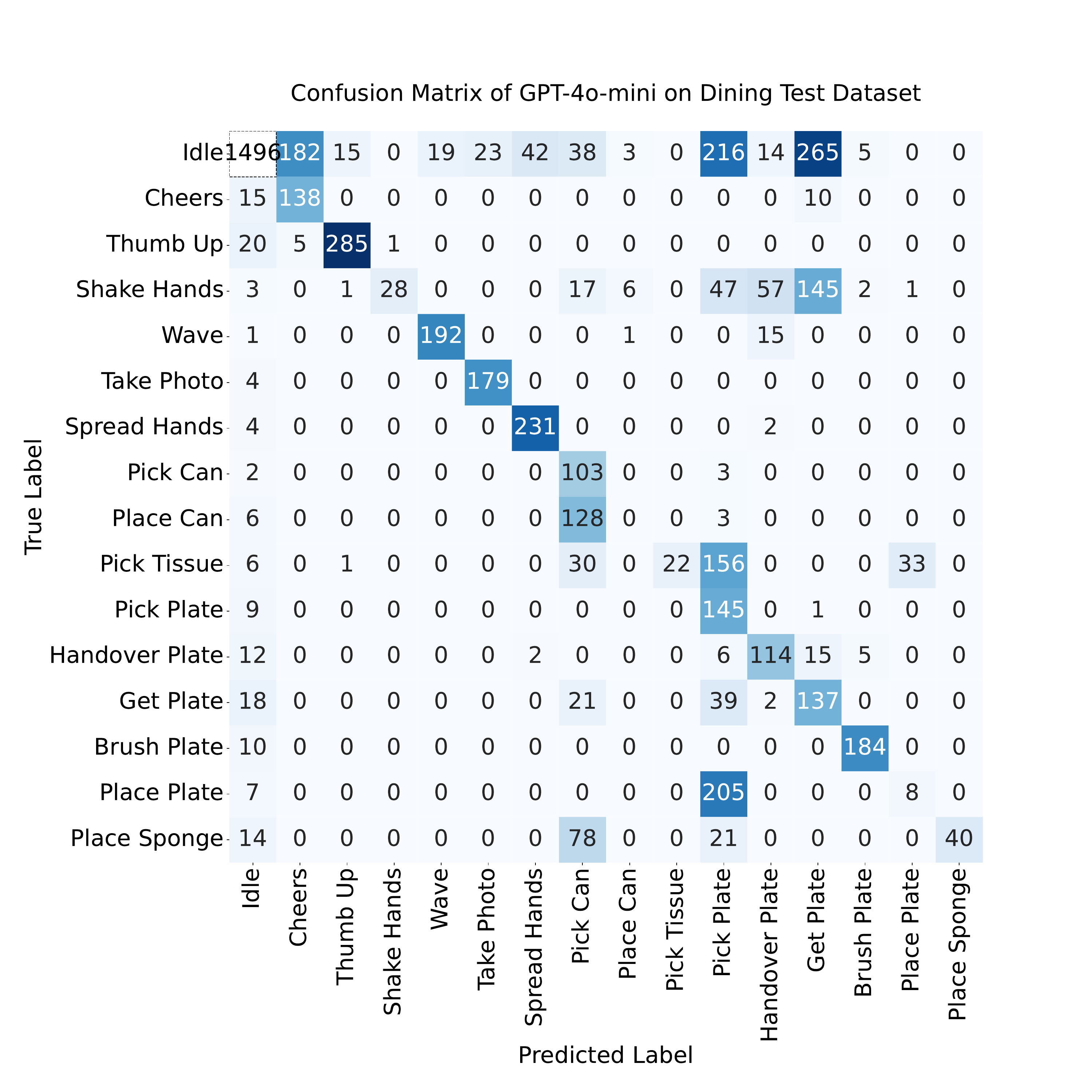}
        \label{fig:image5}
    \end{minipage}
    \begin{minipage}{0.48\linewidth}
        \centering
        \includegraphics[height=8.5cm, keepaspectratio]{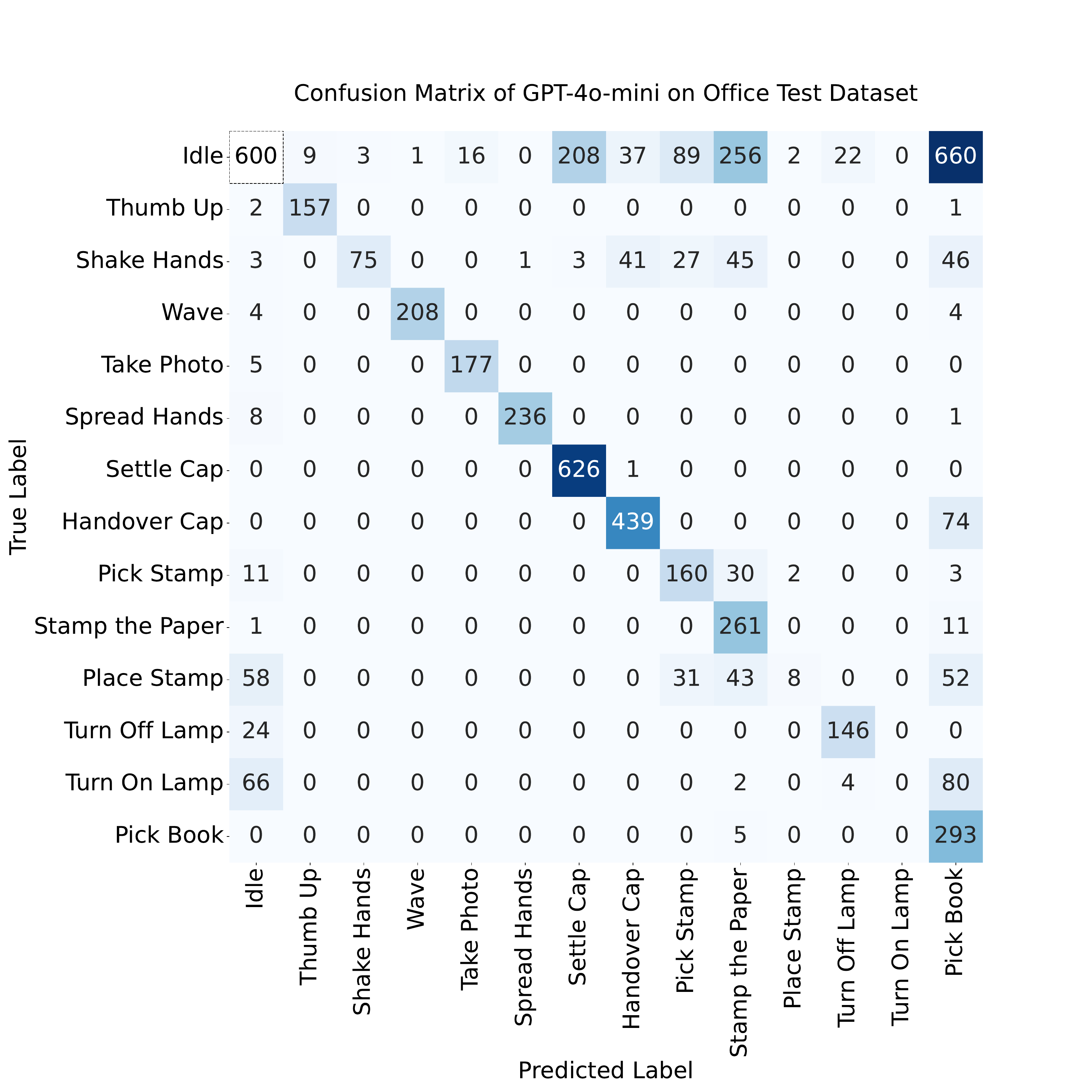}
        \label{fig:image6}
    \end{minipage}
    
    \caption{\textbf{Confusion Matrices of Our Model (with and without Human Details) and GPT-4o-mini.} To show the results more clearly, we did not color the cell in the top left corner since ”idle” accounts for a significant proportion in the data.}
    \label{fig:Confusion Matrix}
\end{figure*}

We use confusion matrices to show the classification performance of our planner on the test dataset. The confusion matrices for our model, our model without human details and GPT-4o-mini on the test datasets of the dining and office scenarios are shown in \Cref{fig:Confusion Matrix}. 

As is shown in the confusion matrices, although the model mainly relies on human body motion and human hand motion input for classification, human details can help the model better deal with certain situations, such as avoiding mis-classification into Idle.

\noindent\textbf{2) Objects Manipulation}

The detailed success rates and average execution times across skills are presented in \Cref{tab:manipulation_detailed}, from which the statistics in \Cref{tab:manipulation} are derived.

In most skills, the manipulation module of \our autonomously executes motions following the patterns of teleoperation data within a comparable time frame.
Trained exclusively on successful human teleoperation cases, the module demonstrates both effectiveness and robustness to slight scene variations during deployment. 
As a result, it achieves higher success rates in skills involving simple motions with abundant training data, such as \textit{Pick Can}, \textit{Handover Plate}, and \textit{Place Stamp}.

However, certain skills pose challenges for the manipulation module. In \textit{Place Plate to Stack} and \textit{Stamp the Paper}, the robot hesitates to drop the plate or press the stamp due to prediction noise. 
In \textit{Pick Plate from Table}, it must overcome increased friction against the table when joint positions deviate from those in the collected data.
Another challenge arises in \textit{Brush with Sponge}, where the success signal predictor struggles to assess the progress of the periodic motion accurately. 
As a result, termination is constrained by a $10$-second timeout. 
These various factors contribute to a longer average execution time for these four skills compared to human performance.

Referring to the experiment on in-skill interruption data presented in \Cref{tab:safety_manipulation}, we select \textit{Pick Can}, \textit{Stamp the Paper}, and \textit{Place Plate to Stack} as representative skills for interruptions occurring during the stages of fetching an object, operating with the object, and returning the object, respectively.

To ensure a controlled data volume across different interruption ratios, we assign a fixed data amount of $M$ to each skill. 
In a full data collection for any given skill, the total data amount is $N$, with $N_{in}$ representing the portion containing in-skill interruptions. 
The ratio of interrupted data in a selected subset is denoted as $\alpha$, meaning that $\lceil\alpha M\rceil$ slices contain interruptions. 
To maintain this proportion, we set $M=N-N_{in}+1$. 
Specifically, $M$ is set to $69$, $66$, and $76$ for the three skills, respectively.

It is worth noting that the assigned data amount is smaller than that used in full data collection models (see \Cref{tab:skills}), which results in a decrease in skill success rate and interrupt success rate compared to the final model.

Each ACT model in this experiment uses the same hyper-parameters as those employed for the corresponding skill in both training and deployment with the full data collection.

\noindent\textbf{3) End2End Policy}

\begin{table}[htb]
    \begin{center}
        \caption{Detailed success rate of \our framework and end-to-end model on different skills.}
        \label{tab:framework_detailed}
        \begin{tabular}{cc|ccccc}
            \toprule
            \multicolumn{2}{c|}{\textbf{Method}} & \textbf{cheers} & \textbf{pick} & \textbf{place} & \textbf{handshake} & \textbf{wave} \\
            \midrule
\multicolumn{2}{c|}{Ours} & 1.00 & 1.00 & 0.85 & 1.00 & 1.00 \\
            \midrule
\multirow{2}{*}{E2E/1} & I.D. & 1.00 & -  & -  & -  & -  \\ & O.O.D.  & 0.95 & -  & -  & -  & -  \\
            \midrule
\multirow{2}{*}{E2E/3} & I.D. & 1.00 & 0.75 & 0.35 & -  & -  \\ & O.O.D.  & 0.95 & 0.30 & 0.45 & -  & -  \\
            \midrule
\multirow{2}{*}{E2E/5} & I.D. & 0.95 & 0.70 & 0.80 & 0.85 & 0.90 \\ & O.O.D.  & 1.00 & 0.55 & 0.35 & 0.60 & 0.85 \\
            \bottomrule
        \end{tabular}
    \end{center}
\end{table}

The detailed result to derive \Cref{tab:framework} is shown in \Cref{tab:framework_detailed}.
For each single skill, we collected $100$ slices of motion in the dataset, which is close to the average volume of all the manipulation skills.
As the small model is not capable of the skills which are totally unseen, we only examine each E2E model with the success rate of skills in the training data.

\end{document}